\documentclass[11pt]{article} 
\usepackage[paperheight=9.44in,paperwidth=6.69in,left=1.85cm,right=1.8cm,bottom=2.5cm,top=3cm,twoside]{geometry}

\usepackage{subfig} 

\usepackage[ruled,vlined,boxed]{algorithm2e}
\usepackage{authblk} 
\usepackage{fancyhdr} 
\usepackage{lastpage} 
\usepackage{textcomp} 
\usepackage{amsmath} 
\usepackage{amssymb} 
\usepackage{amsthm} 
\usepackage{graphicx} 
\usepackage{multirow} 
\usepackage{hyperref} 
\usepackage{longtable} 
\usepackage{float} 
\usepackage{enumitem} 
\usepackage[T1]{fontenc} 
\usepackage[utf8]{inputenc} 

\newcommand{\linia}{\noindent\rule{\linewidth}{0.25mm}\hrulefill} 

\SetAlFnt{\small}
\SetAlCapFnt{\small}
\SetAlCapNameFnt{\small}
\SetAlCapHSkip{0pt}
\IncMargin{-\parindent}

\usepackage{times}
\usepackage{titlesec}

\titleformat*{\section}{\large\bfseries}
\titleformat*{\subsection}{\normalsize\bfseries}

\fancypagestyle{firststyle}
{
	\fancyhf{}
	\setcounter{page}{35} 
	\fancyfoot{}
	\fancyhead{}
	\fancyhead[CE]{\upshape Zeszyty Naukowe WWSI, No 24, Vol. 15, 2021, pp.\thepage-\pageref{LastPage} \newline DOI: 10.26348/znwwsi.24.35}
	\fancyhead[CO]{\upshape Zeszyty Naukowe WWSI, No 24, Vol. 15, 2021, pp. \thepage-\pageref{LastPage} \newline DOI: 10.26348/znwwsi.24.35}
	\fancyfoot[L,R]{\footnotesize\bfseries Manuscript received July 17, 2021}
	\fancyfoot[LE,RO]{\thepage}
	}

\title{\large \bfseries Principal Component Analysis versus Factor Analysis} 
\author{\normalsize Zenon Gniazdowski\thanks{E-mail: zgniazdowski@wwsi.edu.pl}}
\affil{\normalsize Warsaw School of Computer Science}

\date{\vspace{-5ex}}
\providecommand{\keywords}[1]{\textbf{\textit{Keywords ---}} #1}

\begin{document}
	
	\maketitle 
	\thispagestyle{firststyle} 
	
	\linia
	
	\begin{abstract}\label{abstract}
		\noindent The article discusses selected problems related to both principal component analysis (PCA) and factor analysis (FA). In particular, both types of analysis were compared. A vector interpretation for both PCA and FA has also been proposed. The problem of determining the number of principal components in PCA and factors in FA was discussed in detail. A new criterion for determining the number of factors and principal components is discussed, which will allow to present most of the variance of each of the analyzed primary variables. An efficient algorithm for determining the number of factors in FA, which complies with this criterion, was also proposed. This algorithm was adapted to find the number of principal components in PCA. It was also proposed to modify the PCA algorithm using a new method of determining the number of principal components. The obtained results were discussed.
	\end{abstract}
	\label{keywords}
	
	\section{Introduction}
	To be able to talk about factor analysis in the context of principal component analysis, the details of both methods should be compared, starting with the algorithms and ending with the effects of both types of analysis. Only selected problems related to principal component analysis (PCA) and factor analysis (FA) will be discussed in this article. First of all, the common elements of both analyzes will be presented, but also the differences between them. Principal component analysis and factor analysis will be performed for the sample data set. In both analyzes, a matrix of correlation coefficients will be used. Additionally, in the case of factor analysis, considerations will be limited to exploratory factor analysis (EFA) using principal components and Varimax rotation. Also, the elements on the diagonal of the correlation matrix will not be reduced by the value of the common variances.
	
	A detailed comparison of the PCA with the FA will allow conclusions to be drawn about the relationship between both types of analysis. This will allow a broader view of the criteria for determining the number of principal components or factors in both types of analysis. As a consequence, it will enable the development of a new efficient algorithm for determining the number of factors in FA and principal components in PCA.
	
	\section{Preliminaries}
	This section introduces the basic concepts or notations that you will use later in this article. In particular, this applies to some letter symbols, abbreviations, basic statistics, rotation of the coordinate system, criteria for determining the number of factors or principal components, as well as the factor analysis algorithm which is based on principal components. 
	Principal components analysis algorithm will not be presented here. This algorithm has already been presented in the article \cite{Gniazdowski2017}.
	\subsection{Notes on symbols, abbreviations and terms}\label{notesOn}
	The article makes some assumptions about the understanding of symbols, abbreviations, and terms. Due to the possibility of their misinterpretation, the above ambiguities will be explained here:
	\begin{itemize}
		\item First, it is necessary to clarify the meanings of some of the abbreviations that were used in the article.
		It is about explaining the following three abbreviations that appear many times in the article:
		\begin{itemize}
			\item PCA -- Principal Component Analysis,
			\item FA -- Factor Analysis,
			\item PC -- Principal component.
		\end{itemize}
		\item An explanation should also be given regarding the meaning of both the uppercase ''X'' and ''Y'' and the lowercase ''x'' and ''y'' that were used in the article. 	Capital letters ''X'' or ''Y'' have been reserved to denote primary variables that have not been processed in any way. This means that the primary variables were not reduced by the constant component (i.e. by the average value) and were also not standardized.
		On the other hand, lowercase ''x'' and ''y'' have been reserved for standardized variables. 
		
		The exception to the above rule are subsections \ref{randComp} and \ref{Varimax}. In subsection \ref{randComp}  a lowercase letter ''x'' denotes the random component of the primary  variable. In subsection \ref{Varimax}, where the Varimax algorithm is described, lowercase ''x'' and ''y'' were reserved for the variables before rotation, and uppercase ''X'' and ''Y'' were reserved for the variables after rotation.
		
		Therefore, starting from section \ref{PCAvsFA}, apart from the case of presenting basic statistics of primary variables, subsequent random variables will be consistently described with a lowercase ''x''. This is because both PCA (as used in this article) and FA refer to standardized random variables.
		\item In this article, both PCA and FA will use a matrix of correlation coefficients. Therefore, in both types of analysis, it is sufficient to consider standardized primary variables instead of the original primary variables. The assumption about the standardization of primary variables is not a limitation here for three reasons:
		\begin{enumerate}
			\item Standardization of random variables does not affect the matrix of correlation coefficients.
			\item Both types of analysis work on standardized primary variables. FA identifies the linear model of standardized primary variables as a function of independent factors (also standard random variables), while PCA transforms standardized primary variables into independent principal components.
			\item Using the transformation of formula (\ref{r7}), one can find the primary variables $ X $ from the standardized primary variables $ x $.
		\end{enumerate}
		\item The article will examine principal components analysis as well as factor analysis. Certain algorithms exist in both types of analysis. They are either common or analogous. The common algorithm is eigenproblem solving for the matrix of correlation coefficients. However, an analogous algorithm is the algorithm for determining the number of principal components in the principal components analysis, as well as the algorithm for determining the number of factors in the factor analysis. An analogous algorithm is also the rotation of the coordinate system. In principal components analysis, rotation enables the identification of principal components, and in factor analysis, rotation enables the identification of optimal factors.
		
		When discussing analogous algorithms, referring to both factors and principal components, the article will use a conglomerate of two words: ''factor/component''. In the context of principal component analysis, this conglomerate will only refer to the principal components. In the context of factor analysis, this conglomerate will only refer to factors.
	\end{itemize}
	
	\subsection{Definitions of basic concepts}
	The author presents here the definitions of elementary concepts in a minimalistic way, without analyzing them in depth. The relevant formulas will be listed here, which are used later in this article. Among them there will be formulas defining basic statistics, such as mean, variance and standard deviation, but also such definitions of such terms as random component of a random variable, standardized random variable, or independent random variables.
	\subsubsection{The mean value of a random variable}
	A random variable $X$ is considered. In particular, a random sample of cardinality $n$ is available. The elements of $X_i$ represent the $ i-th $ implementation of the random variable. The estimator of the expected value of this random variable is its mean value:
	\begin{equation}\label{r1}
		\overline{X}=\frac{1}{n}\sum_{i=1}^{n}X_i.
	\end{equation}
	\subsubsection{Random component of a random variable}\label{randComp}
	In order to be able to assess the dispersion of variable $X$, its mean value can be subtracted from its individual implementations.
	In this way, the random component $ x $ of the random variable $ X $ is obtained. Individual implementations $x_i$ of the variable $ x $ will take the form:
	\begin{equation}\label{r2}
		x_i=X_i-\overline{X}.
	\end{equation}
	It is a random variable $ X $ reduced by a constant component.
	\subsubsection{Variance of a random variable}
	A measure of the variability of a random variable $ X $ is its variance. It is defined as the mean value of the squares of individual implementations of the random component $ x $ of the random variable $ X $. The variance estimator $ v $ is given by the formula:
	\begin{equation}
		v=\frac{1}{d}\sum_{i=1}^{n}x_i^2.
	\end{equation}
	For $ d = n-1 $ the variance estimator $ v $ is unbiased. For $ d = n $ it is the biased estimator. The biased estimator gives an underestimated result of the variance. The ratio of the biased variance estimator to the unbiased variance estimator is $ \left (n-1 \right) / n $. The limit value of this ratio is:
	\begin{equation}
		\lim_{n\rightarrow\infty}{\left(\frac{n-1}{n}\right)}=1.
	\end{equation}
	This means that as the value of $ n $ increases, the biased estimator follows the unbiased estimator. Therefore, it can be said that for $ d = n $ the variance estimator is an asymptotically unbiased estimator \cite{Francuz2007}. For example, for $ n> 30 $ the estimation error of the biased estimator is less than $ 3.3\% $, and for $  n> 150 $ the error is less than $ 0.67\% $. In practice, $ n $ is usually quite large, so for the purposes of this article, it is assumed that the variance $ v $ will be calculated from the formula:
	\begin{equation}
		v=\frac{1}{n}\sum_{i=1}^{n}x_i^2.\label{r5}
	\end{equation}
	\subsubsection{Standard deviation of a random variable}
	Another measure of the variability of a random variable $ X $ is its standard deviation defined as the square root of the variance $ v $. The estimator of the standard deviation will be denoted as $ s $:
	\begin{equation}\label{r6}
		s=\sqrt{\frac{1}{n}\sum_{i=1}^{n}x_i^2}.
	\end{equation}
	\subsubsection{Standardized random variable}
	If the variable $ X $ is normally distributed with the mean value $ \overline{X} $ and the standard deviation $ s $, then it can be standardized by performing the following transformation:
	\begin{equation}\label{r7}
		x_i:=\frac{X_i - \overline{X}}{s}=\frac{x_i}{s}.
	\end{equation}
	After standardization, the variable $ x $ has the mean value $ \overline{x} = 0 $ and the standard deviation $ s = 1 $.
	\subsubsection{Pearson's correlation coefficient}
	A measure of the relationship between two random variables $ X $ and $ Y $ is their covariance. The normalized covariance to one is called the Pearson correlation coefficient:
	\begin{equation}
		R_{X,Y}=\frac{\sum_{i=1}^{n}\left[\left(X_i-\overline{X}\right)\left(Y_i-\overline{Y}\right)\right]}{\sqrt{\sum_{i=1}^{n}\left(X_i-\overline{X}\right)^2}\sqrt{\sum_{i=1}^{n}\left(Y_i-\overline{Y}\right)^2}}.
	\end{equation}
	Using (\ref{r2}), the formula for the correlation coefficient can be transformed to the form:
	\begin{equation}\label{r9}
		R_{X,Y}=\frac{\sum_{i=1}^{n}{x_iy_i}}{\sqrt{\sum_{i=1}^{n}x_i^2}\sqrt{\sum_{i=1}^{n}y_i^2}}.
	\end{equation}
	In the numerator of the formula there is the dot product of two vectors $ x $ and $ y $, and in the denominator there is the product of the lengths of these vectors. This means that the correlation coefficient is identical to the cosine of the angle between the two random vectors $ x $ and $ y $ \cite{Gniazdowski2013}:
	\begin{equation}\label{r10}
		R_{X,Y}=\frac{x\cdot y}{\left\| x\right\| \cdot \left\| y\right\| }=\cos{(x,y)}.
	\end{equation}
	Here it should also be added that the standardization of a random variable does not change the correlation coefficient.
	\subsubsection{Determination coefficient}
	The coefficient of determination describes the level of common variance of two correlated standardized random variables. This coefficient is a good measure to describe the similarity between the correlated random variables \cite{Gniazdowski2017}.
	\subsubsection{Independent random variables}
	If the random variables are independent, then the estimate of the correlation coefficient between these variables is close to zero. Also, the value of the coefficient of determination, measuring the level of common variance, describing the level of similarity between the two random variables is close to zero. Here it should be noted that if the random variables are independent, then the variance of the sum of these variables is equal to the sum of their variances \cite{Francuz2007}.
	\subsection{Rotation of the coordinate system}
	Given is an n-dimensional space with a Cartesian orthogonal coordinate system. In this space there is a point whose coordinates define the n-element vector. The axes of a given coordinate system are rotated around the center so that after rotation the system is still an orthogonal system. In the new coordinate system (after rotation), the point will not change its position, but will acquire new coordinates, i.e. the vector defining the point's location will change. To solve the problem of changes in vector components, the transformation of the coordinate system should be described. This transformation is described by the orthogonal matrix $ R $. Its elements $ R_{ij} $ are the cosines of the angles between the $ i-th $ axis of the new coordinate system and the $ j-th $ axis of the old coordinate system (row number identifies the new axis, column number identifies the old axis \cite{Gniazdowski2017}). It is assumed that:
	\begin{itemize}
		\item The old coordinate system is the standard system. Its successive axes are successive standard unit vectors that constitute the identity matrix.
		\item The new coordinate system is an orthogonal system whose axes are described in the old coordinate system as vectors of unit length. Successive columns of these vectors will form an orthogonal matrix $ U $.
	\end{itemize}
	In such cases, the rotation matrix from the standard coordinate system to the final coordinate system is a matrix whose rows are the direction vectors of the axis of the final coordinate system \cite{Gniazdowski2017}:
	\begin{equation}\label{r11}
		R=U^T.
	\end{equation}
	\subsubsection{Vector rotation}
	The rotation matrix $ R $ is used to find the coordinates of the vector $ v $ in the new coordinate system that arose with the orthogonal rotation of the axes of the old coordinate system. The vector $ v=\left[v_1,\ldots,v\right]^T $ in the new coordinate system will receive new components $v^\prime$:
	\begin{equation}\label{r12}
		v^\prime:=R v.
	\end{equation}
	When the vector $ v $ specifies one point in space, it is represented as a row in some matrix. This means that its transposition is available. Also the resulting vector will be a row in the matrix. In order to rotate the vector represented in this way, both sides of the formula (\ref{r12}) should be transposed:
	\begin{equation}
		\left(v^\prime\right)^T=(Rv)^T.
	\end{equation}
	The result is a formula to rotate the row vector:
	\begin{equation}\label{r14}
		\left(v^\prime\right)^T=v^T R^T = v^T U.
	\end{equation}
	If instead of single points $ v^T $ and $ (v^\prime)^T $ we consider sets of points in the form of rectangular matrices $ M $ and $ M^\prime $, in which vectors $ v^T $ and $ (v^\prime)^T $ will be single rows, then equation (\ref{r14}) becomes the matrix equation:
	\begin{equation}\label{r15}
		M^\prime=M U.
	\end{equation}
	In this way, the rows of the factor loadings matrix are rotated in factor analysis, and the standardized primary variables are transformed into principal components in principal components analysis.
	\subsubsection{Rotation on the plane}
	The n-dimensional space is considered. In this space, the orthogonal Cartesian coordinate system is considered. This system is defined by $ n $ orthogonal axes $ X_1, X_2,\ldots, X_n $. In the above space, any pair of different $ X_i $ and $ X_j $ axes (i.e. such that $ i \neq j $) define a plane. In a given plane, rotation by angle $ \varphi $ means simultaneous rotation of both axes. As the position of both axes changes, the coordinates of the points will also change. Usually the aim is to get to know the new coordinates of the points in the new coordinate system. For this purpose, the orthogonal rotation matrix $ R $ is calculated, and then new coordinates of the points are calculated using it.
	
	In order to rotate the axis by a given angle $ \varphi $ on a given  $ \left(X_i, X_j \right) $ plane, the rotation matrix $ r_{ij} $ must be built for this angle. Denoting $ \cos{\varphi} $ as $ c $ and $ \sin{\varphi} $ as $ s $, the rotation matrix in the $ \left(X_i, X_j \right) $ plane is a modified identity matrix with four elements changed: $ r_{ii} = c $, $ r_{ij} = s $, $ r_{ji} = - s $, $ r_{jj} = c $ \cite{Legras1974}:
	\begin{equation}
		r_{ij}=\left[\begin{matrix}\begin{matrix}1&\cdots&0\\\vdots&\ddots&\vdots\\0&\cdots&1\\\end{matrix}&\begin{matrix}&&\\&&\\&&\\\end{matrix}&\begin{matrix}&&0\\&&\\&&\\\end{matrix}\\\begin{matrix}&&\\&&\\&&\\\end{matrix}&\begin{matrix}c&&s\\&\begin{matrix}1&\cdots&0\\\vdots&\ddots&\vdots\\0&\cdots&1\\\end{matrix}&\\-s&&c\\\end{matrix}&\begin{matrix}&&\\&&\\&&\\\end{matrix}\\\begin{matrix}&&\\&&\\0&&\\\end{matrix}&\begin{matrix}&&\\&&\\&&\\\end{matrix}&\begin{matrix}1&\cdots&0\\\vdots&\ddots&\vdots\\0&\cdots&1\\\end{matrix}\\\end{matrix}\right]
	\end{equation}
	\subsubsection{Composition of rotations}
	If there is a need to make successive rotations on all $ \left(X_i,X_j\right) $ planes defined by different pairs of axes $ \left(X_i\neq X_j\right) $, then the resultant rotation matrix $ R $ is the product of successive rotation matrices:
	\begin{equation}
		R = \prod_{^{i=1,2,\ldots,n-1}_{j=i+1,\ldots,n}}  r_{ij}.
	\end{equation}
	The algorithm for finding the final matrix describing the resultant rotation is as follows:
	\begin{enumerate}
		\item $ R:= I; $
		\item $ \forall_{i,j(^{i=1,2,\ldots,n-1}_{j=i+1,\ldots,n})}\ R:=R\cdot r_{ij} $.
	\end{enumerate}
	Since the matrix $ r_{ij} $ describing rotation in a given plane is a modified identity matrix with changed four elements, therefore in the second point of the algorithm there is no need to fully multiply the matrices $ R $ and $ r_{ij} $, but it is enough to modify the elements of the matrix $ R $ in the $ i- $th and the $ j- $th rows, as well as in the $ i- $th and $ j- $th columns. First, the elements $ R_{ii}^\prime $ and $ R_{jj}^\prime $ on the diagonal can be found, as well as the non-diagonal elements $ R_{ij}^\prime $ and $ R_{ji}^\prime $:
	\begin{equation}
		\begin{split}
			R_{ii}^\prime :=& c^2r_{ii}+sc\left(r_{ij}+r_{ji}\right)+s^2 r_{jj};\\
			R_{jj}^\prime :=& -c^2r_{jj}+sc\left(r_{ij}+r_{ji}\right)-s^2 r_{ii};\\
			R_{ij}^\prime :=& sc\left(r_{jj}-r_{ii}\right)+c^2 r_{ij}-s^2 r_{ji};\\
			R_{ji}^\prime :=& scr_{jj}-r_{ii}+c^2 r_{ji}-s^2 r_{ij}.
		\end{split}
	\end{equation}
	Also, the remaining elements in both rows and both columns should be calculated:
	\begin{equation}
		\left.
		\begin{split}
			R_{ip}'&:=cR_{ip}+sR_{jp}\\
			R_{jp}'&:=-sR_{ip}+cR_{jp}\\
			R_{pi}'&:=cR_{pi}+sR_{pj}\\
			R_{pj}'&:=-sR_{pi}+cR_{pj}
		\end{split}
		\right\rbrace   p\ne i,p\ne j.
	\end{equation}
	\subsection{Criteria for determining the appropriate number of principal components or factors}\label{selectCommon}
	In principal component analysis as well as in factor analysis, it is important to determine the appropriate number of principal components and factors. Their number should at least allow for sufficient representation or modeling of the primary variables. For this purpose, various criteria are used \cite{Larose2006,Mooi2018,Thompson2004}: the criterion of the scree plot, the criterion of the total variance explained by the factors/components, the Kaiser criterion or the unit eigenvalue criterion, the criterion of the number factors/components not greater than half the number of the primary variables. These criteria will be discussed in more detail:
	\begin{itemize}
		\item Scree Plot Criterion: This is a graphical method in which the plot shows successive eigenvalues, from largest to smallest. The shape of the graph resembles a scree. Usually, at the beginning, the curve drops sharply, and in the later part, behind the so-called ''elbow'', it descends more gently. As many factors/components are taken as the eigenvalues are located on the slope of the scree. In practice, it is assumed that there are as many factors/components as the eigenvalues are above the ''elbow'' of the scree.
		\item Percentage criterion of the explained variance: It is assumed that there are so many factors/components that the sum of the eigenvalues associated with the successive factor/component is not less than the established percentage threshold related to the trace of the correlation matrix.
		\item Eigenvalue criterion called the Kaiser criterion: It is assumed that there should be as many factors/components as the eigenvalues of the correlation matrix are not less than one. A single standardized variable has a variance of one. Any factor/component with an eigenvalue greater than one accounts for more variance than a single variable. The rationale for using the eigenvalue criterion is that each factor/component should represent or explain at least one primary variable. That is, only factors/components with eigenvalues not less than one should be kept. Since the goal of both PCA and FA is to reduce the total number of factors/components, each factor/component should account for a greater variance than the variance of a single primary variable.
		\item The criterion of half the number of primary variables: It is assumed that the number of factors/components should not exceed half the number of all primary variables. If the identification of principal components is treated as lossy compression, it is important that this compression significantly reduces the size of the stored set. A file that is half the size of the original file can be considered sufficiently compressed. On the other hand, the number of factors/components in the factor/component model, not greater than half of the primary variables (including potentially possible factors/components), is satisfactory from the point of view of the simplicity of the model.
	\end{itemize}
	It should be emphasized at this point that none of the above criteria should be regarded as absolute criteria, but rather as subsidiary criteria. First of all, it may happen that particular criteria may produce different or inconclusive results:
	\begin{itemize}
		\item The scree criterion may not apply as the two phases clearly separated by a so-called ''elbow''' may not be visible in the scree plot.
		\item The percentage criterion of the explained variance may also give unsatisfactory results, despite the relatively large variance represented. The number of factors/components resulting from this criterion may be too small to adequately represent the primary variables.
		\item The Kaiser criterion can also falsify the number of factors/components. For example, for data describing the petals of an iris flower \cite{Fisher1936}, the second eigenvalue for the correlation coefficient matrix is less than one. Nevertheless, only two factors/components can satisfactorily represent the primary variables \cite{Gniazdowski2017}.
		\item Also, the criterion of half the number of primary variables may be too strict. For example, according to this criterion, with an odd number of variables, the number of factors/components should not be greater than $ (n-1)/2 $. Meanwhile, it would be better if this number was $ (n+1)/2 $.
	\end{itemize}
	In such an ambiguous situation, the number of factors/components should be decided by analyzing the full context of the study.
	\subsection{Factor analysis algorithm}
	In factor analysis, on the basis of the analysis of the set of observed correlated random variables, linear models of these variables are built with respect to the set of factors that are independent random variables with unit variance. Factors are subject to interpretation. If the interpretation of the factors is made, then through these interpreted factors conclusions can be drawn about the causes of the variability of the observed variables. Various approaches to FA are mentioned in the literature. These approaches relate to the types of FA, algorithms used, as well as rotation methods (e.g. \cite{Mooi2018,Thompson2004,Ertel2013,Hofacker2007,Phakiti2018}:
	\begin{itemize}
		\item There are two types of factor analysis. The first type is exploratory factor analysis (EFA), the second is confirmatory factor analysis (CFA). In exploratory factor analysis, neither a priori relationship between the observed variables and factors nor the number of factors is assumed. On the other hand, confirmatory factor analysis presupposes some knowledge of the model, which may be confirmed during the course of it. In practice, some researchers may run EFA first and then use CFA to validate or confirm EFA results. In turn, EFA is not a necessary condition for the CFA \cite{Phakiti2018}.
		\item In FA, in the context of the algorithms used, there are at least two different approaches: the principal component approach and the maximum likelihood approach.
		\item The factors obtained from the principal components need not be final factors. Factors can be rotated for simpler interpretation. The main division of rotation methods is between orthogonal and oblique rotation. Different possible rotation criteria lead to different possible rotation methods. Hence, at least the following methods of factor rotation are known: Varimax - orthogonal rotation, Quartimax - orthogonal rotation, and Oblimin - oblique rotation.
	\end{itemize}
	Only some aspects of the factor analysis will be presented in this article:
	\begin{itemize}
		\item Exploratory Factor Analysis,
		\item Method based on principal components,
		\item Varimax rotation.
	\end{itemize}
	Since in factor analysis standardized primary variables are modeled as a function of independent factors, therefore, without losing generality in the remainder of this article, all mathematical formulas describing factor analysis will only apply to standardized primary variables.
	
	Having a set of $ n $ random variables $ x=[x_1,\ldots,x_n] $, we can proceed to FA.
	In the $ n- $dimensional space defined by these variables, $ m $ measurement points are considered. The data is stored in the form of the $ x_{m\times n} $ matrix:
	\begin{equation}\label{r20}
		x=\left[\begin{matrix}x_{11}&\cdots&x_{1n}\\
			\vdots&\vdots&\vdots\\x_{m1}&\cdots&x_{mn}\\
		\end{matrix}\right].
	\end{equation}
	The individual columns of the $ x $ matrix contain $ n $ successive $ x_i $ random variables. Each $ i- $th random variable creates a column random vector $ x_i\ \left(i=1,2,\ldots,m\right) $:
	\begin{equation}
		x_i=[x_{1i},x_{2i},\ldots\ ,x_{mi}]^T
	\end{equation}
	In turn, the $ j- $th row of the $ X $ matrix represents a single measurement point $ p_j $, containing the $ j- $th elements of successive random variables $ x_i\ (i=1,2,\ldots,n) $:
	\begin{equation}
		p_j=[x_{j1},x_{j2},\ldots\ ,x_{jn}].
	\end{equation}
	\subsubsection{Steps of the factor analysis algorithm}
	\begin{enumerate}
		\item For all n random variables stored in the matrix $ x $, the matrix of correlation coefficients $ R $ is calculated:
		\begin{equation}
			R=\left[\begin{matrix}1&\cdots&R_{1n}\\
				\vdots&\ddots&\vdots\\
				R_{n1}&\cdots&1\\\end{matrix}\right]
		\end{equation}
		Its components are $ R_{ij} $ elements, which are the correlation coefficients (\ref{r9}) between all the $ x_i $ and $ x_j $ variables.
		\item For the matrix of correlation coefficients $ R $, it is necessary to solve the eigenproblem. As a result, a diagonal matrix $ \Lambda $ is obtained containing on the diagonal sorted non-increasing successive eigenvalues of $ \lambda_i $, representing the variances of potential factors:
		\begin{equation}\label{r24}
			\Lambda=\left[\begin{matrix}\lambda_1&\cdots&0\\\vdots&\ddots&\vdots\\0&\cdots&\lambda_n\\\end{matrix}\right].
		\end{equation}
		\item The matrix $ U $ is also obtained, which in its columns contains successive eigenvectors corresponding to the successive eigenvalues:
		\begin{equation}\label{r25}
			U=\left[\begin{matrix}U_{11}&\cdots&U_{1n}\\
				\vdots&\ddots&\vdots\\U_{n1}&\cdots&U_{nn}\\
			\end{matrix}\right].
		\end{equation}
		\item From the $ \Lambda $ matrix, a diagonal $ S $ matrix containing standard deviations of potential factors is calculated:
		\begin{equation}\label{r26}
			S=\sqrt\mathrm{\Lambda}=\left[\begin{matrix}\sqrt{\lambda_1}&\cdots&0\\
				\vdots&\ddots&\vdots\\0&\cdots&\sqrt{\lambda_n}\\
			\end{matrix}\right].
		\end{equation}
		\item Using the matrices $ U $ and $ S $ there is a square matrix of factor loadings $ L $:
		\begin{equation}\label{r27}
			L=U\cdot S=\left[\begin{matrix}L_{11}&\cdots&L_{1n}\\
				\vdots&\ddots&\vdots\\L_{n1}&\cdots&L_{nn}\\
			\end{matrix}\right].
		\end{equation}
		\item If we assume that $ F=[f_1,\ldots,f_2]^T $ is the set of independent standardized random variables called factors, then the set of standardized $ x=[x_1,\ldots,x_n]^T $ variables can now be represented as a linear model with respect to the set of independent factors:
		\begin{equation}
			\left[\begin{matrix}x_1\\\vdots\\x_n\\
			\end{matrix}\right]=\left[\begin{matrix}L_{11}&\cdots&L_{1n}\\
				\vdots&\ddots&\vdots\\L_{n1}&\cdots&L_{nn}\\
			\end{matrix}\right]\cdot\left[\begin{matrix}f_1\\\vdots\\f_n\\
			\end{matrix}\right].
		\end{equation}
	\end{enumerate}
	\subsubsection{Factors reduction in the factor model}
	Since the factors $ f_1,\ldots,f_n $ are standardized independent random variables, the squares of the $ L_{ij} $ elements contained in the $ L $ matrix represent the variances that are contributed by the individual independent factors $ f_i $ to the individual random variables $ x_1,\ldots,x_n $. In the $ \Lambda $ matrix, the individual eigenvalues are sorted non-ascending. Therefore, the influence of the first factors on the variables $ x_i $ dominates over the last ones. Therefore, the variables $ x_i $ can be made dependent on the first $ k $ dominant factors, ignoring the insignificant factors.
	
	The influence of the omitted factors can be presented as the error vector $ E=[\varepsilon_1,\ldots,\varepsilon_n]^T $, which represents the uncontrollable but also independent of the factors $ f_j $ disturbances, having the nature of random errors:
	\begin{equation}\label{r29}
		\left[\begin{matrix}x_1\\\vdots\\x_n\\
		\end{matrix}\right]=\left[\begin{matrix}L_{11}&\cdots&L_{1k}\\
			\vdots&\ddots&\vdots\\L_{n1}&\cdots&L_{nk}\\
		\end{matrix}\right]\cdot \left[\begin{matrix}f_1\\
			\vdots\\f_k\\\end{matrix}\right]+\left[\begin{matrix}\varepsilon_1\\
			\vdots\\\varepsilon_n\\\end{matrix}\right].
	\end{equation}
	The square matrix of factor loadings $ L $ (\ref{r27}), which originally had the size $ n\times n $, was reduced in formula (\ref{r29}) to a rectangular matrix of size $ n\times k\ (k<n) $. In the $ L $ matrix the number of rows remained the same, and the number of columns decreased:
	\begin{equation}\label{r30}
		L=\left[\begin{matrix}L_{11}&\cdots&L_{1k}\\
			\vdots&\ddots&\vdots\\L_{n1}&\cdots&L_{nk}\\
		\end{matrix}\right].
	\end{equation}
	Expression (\ref{r29}) is a model of the primary variables $ x_i\ (i=1,\ldots,n) $ with respect to the independent factors $ f_j\ (j=1,\ldots,k) $ and the independent vector $ \varepsilon_i\ (i=1,\ldots,n) $. Both the primary variables $ x_i $ and the factors $ f_j $ are standardized random variables with a unit variance (and therefore also with a unit standard deviation).
	
	Each primary variable $ x_i $ is the sum of the random variables derived from the factors and from the error. The variance contributed to a given variable $ x_i $ by the factor $ f_j $ is $ L_{ij}^2 $. Since the factors are independent random variables, the variance $ v_i $ contributed to the variable $ x_i $ by $ k $ factors $ f_j\ (j=1,\ldots,k) $ is equal to the sum of the variances contributed by these factors:
	\begin{equation}
		v_i=\sum_{j=1}^{k}L_{ij}^2.
	\end{equation}
	For each primary variable, the value of $ v_i $ determines the level of variance reproduced by using $ k $ factors in the model (\ref{r29}). The components $ v_i $ form the vector of common variances $ V $:
	\begin{equation}\label{r32}
		V=\left[\begin{matrix}v_1\\\vdots\\v_n\\\end{matrix}\right].
	\end{equation}
	The diversity of the elements of $ V_i $ in the vector $ V $ shows that individual variables are modeled with different accuracy by a selected set of factors. A reasonable model should represent most of the variance of the modeled primary variable. Most means at least $ 50\% $. This level of explaining the variance of the primary variable can be found in \cite{Larose2006}. If the condition $ v_i\le 0.5 $ holds for any $i$, it is information that too few factors were used to explain the primary variables. One or more of the criteria described in Section \ref{selectCommon} may be used to determine the appropriate number of factors.
	\subsubsection{Modeling of primary variables}
	Expression (\ref{r29}) is sufficient to explain the influence of latent factors. It will allow to determine which variable to what extent depends on a given factor. More precisely, the analysis of the matrix (\ref{r30}) is sufficient to explain the influence of latent factors on the primary variables:
	\begin{itemize}
		\item By estimating the vector $ V $ (\ref{r32}) with it, it is possible to obtain information about the level of variance of primary variables $ x_i $ explained by selected factors.
		\item By rotating the factor loadings contained in the rows of this matrix, it is also possible to improve the efficiency of the factor interpretation.
	\end{itemize}
	Unfortunately, the model (\ref{r29}) is not sufficient for a reliable simulation of $ x_i $ variables. This model does not take into account the influence of random disturbances on the primary variables. To get rid of this deficit, model (\ref{r29}) should be extended with a component that will describe the random disturbance vector $ E $.
	
	Vector $ E $ is influenced by omitted factors $ f_{k+1}\ldots f_n $. They can be replaced by one independent unique factor $ f_0 $. The standardized variable $ x_i $ has a unit variance. The component $ v_i $ of the vector $ V $ (\ref{r32}) describes that part of the variance that is explained by $ k $ factors. Since all the selected factors and the factor $ f_0 $ are independent random variables, the factor $ f_0 $ should contribute enough variance to the variable $ x_i $ so that its total variance (both that derived from the independent factors $ f_1,\ldots,f_k $ and that derived from the independent factor $ f_0 $) sums up to ones. Therefore, the error $ e_i $ of the variable $ x_i $ can be expressed in the following form:
	\begin{equation}
		\varepsilon_i=\sqrt{1-v_i}\cdot f_0.
	\end{equation}
	Denoting by $ w_i $ the standard deviation $ \sqrt{1-v_i} $ of the error $ \varepsilon_i $, the error vector $ E $ takes the following form:
	\begin{equation}
		E=\left[\begin{matrix}\varepsilon_1\\
			\vdots\\\varepsilon_n\\
		\end{matrix}\right]=\left[\begin{matrix}w_1\\
			\vdots\\w_n\\
		\end{matrix}\right]\cdot f_0.
	\end{equation}
	In this way, the factor model suitable for simulating the influence of factors on the primary variables, also considering the influence on the primary variables of uncontrolled random disturbances, takes the final form:
	\begin{equation}
		\left[\begin{matrix}x_1\\
			\vdots\\x_n\\
		\end{matrix}\right]=\left[\begin{matrix}L_{11}&\cdots&L_{1k}\\
			\vdots&\ddots&\vdots\\L_{n1}&\cdots&L_{nk}\\
		\end{matrix}\right]\cdot\left[\begin{matrix}f_1\\
			\vdots\\f_k\\
		\end{matrix}\right]+\left[\begin{matrix}w_1\\
			\vdots\\w_n\\
		\end{matrix}\right]\cdot f_0.
	\end{equation}
	In this model, each primary variable is linearly dependent on at least one common factor $ f_i\ (i=1,\ldots,k) $ and on one specific (unique) factor $ f_0 $.
	\subsubsection{Varimax rotation}\label{Varimax}
	The Kaiser-developed Varimax rotation procedure \cite{Kaiser1958} is probably the most popular rotation method. It aims to lead to a simple solution in FA. For Varimax rotation, a simple solution means that each factor has a small number of large factor loadings and a large number of zero (or small) factor loadings. This simplifies the interpretation because after Varimax rotation, each primary variable is usually associated with one or at most a small number of factors. Formally, Varimax looks for the rotation of the initial factors so that the variance of the factor loadings is maximized \cite{Abdi2003}. Later in this subsection, the Varimax rotation procedure will be based on the original Kaiser article \cite{Kaiser1958}.
	
	The idea behind Varimax rotation is that the factor loadings are optimized separately on each of the planes defined by a pair of coordinate axes. If rotation on a given plane does not increase the value of the objective function, then this rotation is ignored by moving to the next plane. Before starting the rotation procedure, the row vectors of the factor loadings matrix are normalized to the unit length. After the rotation is complete, the rotated vectors will be restored to their original length \cite{Kaiser1958}.
	
\begin{figure}[t]
		\centering
		\includegraphics[width=0.65\textwidth]{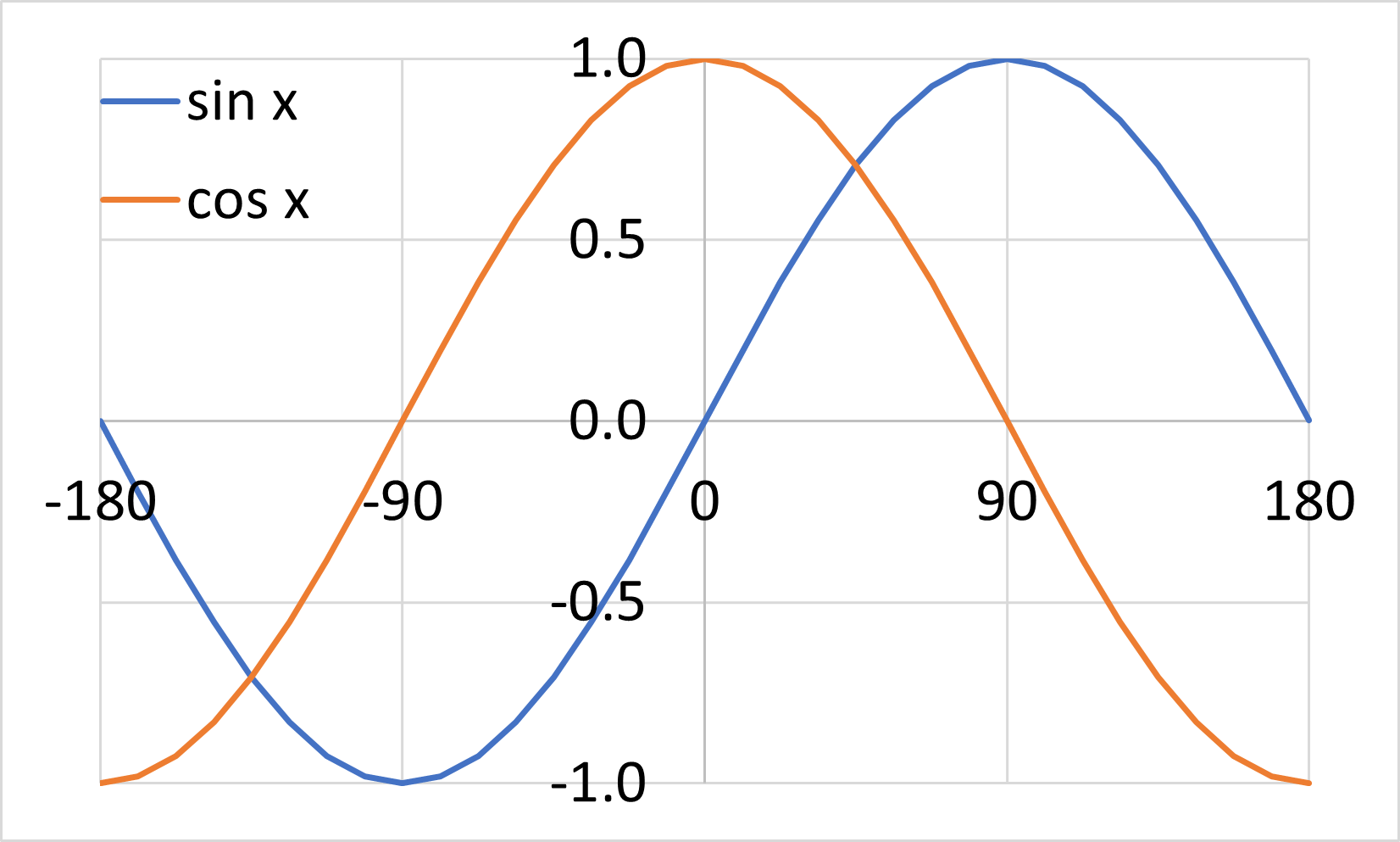}
		\caption{Waveforms of sine and cosine functions}\label{fig1}
	\end{figure}

	Two columns in the factor loadings matrix are considered, which will form a matrix of size $  n\times 2 $. This matrix represents a set of points on the plane labeled $ OXY $. These points were created by projecting the vectors representing the primary variables to the $ OXY $ plane. These points are a two-dimensional representation of these primary variables. The $ i-th $ row of the matrix successively represents the $ i-th $ primary variable in the form of a vector with coordinates $ (x_i, y_i) $. An orthogonal rotation matrix $ R $ is also given, which on the OXY plane describes the rotation of the coordinate system by a given angle $ \phi $:
	\begin{equation}
		R=\left[\begin{matrix}\cos{\phi}&\sin{\phi}\\-\sin{\phi}&\cos{\phi}\\\end{matrix}\right]
	\end{equation}
	Its transposition $ R^T $ will be used to transform the points, i.e. to rotate row vectors:
	\begin{equation}
		R^T=\left[\begin{matrix}\cos{\phi}&-\sin{\phi}\\\sin{\phi}&\cos{\phi}\\\end{matrix}\right].
	\end{equation}
	After the coordinate system is rotated in the new coordinate system, the point with the coordinates $ (x_i, y_i) $ becomes the point with the new coordinates $ (X_i, Y_i) $. Coordinate transformation can be described by matrix multiplication:
	\begin{equation}
		\left[\begin{matrix}X_1&Y_1\\
			\vdots&\vdots\\
			X_N&Y_N\\
		\end{matrix}\right]:=
		\left[ \begin{matrix}
			x_1&y_1 \\
			\vdots & \vdots \\
			x_n&y_n
		\end{matrix}   \right] \cdot 
		\left[\begin{matrix}
			\cos{\phi}&-\sin{\phi} \\
			\sin{\phi}&\cos{\phi}
		\end{matrix}\right]
		.
	\end{equation}
	For the $ i-th $ point ($ i-th $ row in the matrices) we get two equations:
	\begin{equation}\label{r39}
		\left\{
		\begin{split}
			X_i&:=x_i \cos{\phi} + y_i \sin{\phi},\\
			Y_i&:=-x_i \sin{\phi}+y_i\cos{\phi}.
		\end{split}\right.
	\end{equation}
	Hence:
	\begin{equation}\label{r40}
		\left\{
		\begin{split}
			dX_i/d\phi&=Y_i,\\
			dY_i/d\phi&=-X_i.
		\end{split}
		\right.
	\end{equation}
	The maximized objective function has the form:
	\begin{equation}\label{r41}
		n^2v_{xy}=n\sum\left(X^2\right)^2-\left(\sum X^2\right)^2+n\sum\left(Y^2\right)^2-\left(\sum Y^2\right)^2.
	\end{equation}
	Using the equation (\ref{r40}), the objective function (\ref{r41}) can be differentiated with respect to the angle $ \phi $ and after differentiation it can be compared to zero:
	\begin{equation}
		n\sum X Y\left(X^2-Y^2\right)-\sum X Y\sum\left(X^2-Y^2\right)=0.
	\end{equation}
	To solve the problem in the space of variables before rotation ($ x $ and $ y $), the formula (\ref{r39}) should be used. After transformation, the relationship describing the angle of rotation on the OXY plane is obtained:
	\begin{equation}
		4\phi=arctan\frac{2\left[n\sum\left(x^2-y^2\right)\left(2xy\right)-\sum{\left(x^2-y^2\right)\sum\left(2xy\right)}\right]}{n\left\{\sum\left[\left(x^2-y^2\right)^2-2xy^2\right]\right\}-\left\{\left[\sum\left(x^2-y^2\right)\right]^2-\left[\sum\left(2xy\right)\right]^2\right\}}.
	\end{equation}
	If we substitute $ u_i = x_i^2-y_i^2 $ and $ v_i = 2x_i y_i $, then the above expression reduces to a simpler form:
	\begin{equation}\label{r44}
		4\phi=arctan\frac{2\left[n\sum{u_iv_i}-\sum{u_i\sum v_i}\right]}{n\sum\left(u_i^2-v_i^2\right)-\left[\left(\sum u_i\right)^2-\left(\sum v_i\right)^2\right]}.
	\end{equation}
	In the range of full rotation from $ -180^0 $ to $ +180^0 $, the functions $ \sin{4\phi} $ and $ \cos{4\phi} $ reach both negative and positive values (Figure 1). Therefore, the expression $ \arctan{\left(\cdot\right)} $ is ambiguous. As a result of examining the signs of the first and second derivative of the numerator and the denominator in the expression (\ref{r44}), Kaiser's work \cite{Kaiser1958} presents ranges of the angle $ \phi $ depending on the signs of the numerator and the denominator of this expression. Table 1 shows the ranges of the $ 4\phi $ angle values. These ranges are consistent with the ranges of variability of the $ sin $ and $ cos $ functions presented in Figure \ref{fig1}.
	\begin{table}[th!]
		\centering
		\caption{The relationship of the solution of the equation (\ref{r44}) with the signs of its numerator and denominator}\label{tab1}
		\fontsize{8}{12}\selectfont{
			\begin{tabular}{c|c|c|c}
				\cline{3-4}
				\multicolumn{2}{c|}{\ } & \multicolumn{2}{|c}{$ Numerator{\ } sign $} \\ \cline{3-4}
				\multicolumn{2}{c|}{\ } &      $+$       &$-$\\ \hline
				$ Denominator $&$+$   &    $0^0$    to $90^0$&    $-90^0$    to $0^0$\\ \cline{2-4}
				$ sign $&$-$ &    $90^0$    to $180^0$&    $-180^0$    to $-90^0$\\ \hline
		\end{tabular}}
	\end{table}
	
	\section{Principal Component Analysis vs. Factor Analysis}\label{PCAvsFA}
	In order to be able to talk about factor analysis in the context of principal components analysis, both types of analysis should be compared. For the sample dataset, PCA will be performed first, then FA. This will allow conclusions to be drawn about the specific relationships between PCA and FA.
	\subsection{Analyzed data: weather information (Dataset No. 1)}
	The data set containing $ 7 $ random variables was used for the analysis. The set consists of $ 49~987 $ data records that were measured at different times of the day in many weather stations from January 1, 2000 to September 20, 2018. In the further part of the presented analysis, subsequent variables will be marked with symbols $ X_1,\cdots X_7 $. The content of individual variables is interpreted as follows:
	\begin{itemize}
		\item $ X_1 $ -- Sea-level pressure in millibar ($ mbar $);
		\item $ X_2 $ -- Air temperature in degree Celsius ($ ^{0}C $);
		\item $ X_3 $ -- Dew point temperature in degree Celsius ($ ^{0}C $);
		\item $ X_4 $ -- Wind direction in degrees of arc  ($ ^0 $ -- the degree symbol);
		\item $ X_5 $ -- Wind speed in meters per second ($ m/sec $);
		\item $ X_6 $ -- Visibility in metres ($ m $);
		\item $ X_7 $ -- Time of measurement - it is a number in the interval $ [0,1) $. The left endpoint is closed, the right endpoint is open. The lower limit is $ 00:00 $ and the upper limit is $ 24:00 $.
	\end{itemize}
	Basic statistics were estimated for all seven variables. The mean values of the variables, their medians and modes were adopted as the measures of the location of random variable distributions. Standard deviations for all variables as well as their minima and maxima were assumed as measures of dispersion. The results are presented in Table \ref{tab2}. The matrix of correlation coefficients (Table \ref{tab3}) and the matrix of determination coefficients (Table \ref{tab4}) were also estimated for all seven variables. The coefficient of determination (equal to the square of the correlation coefficient) defines the degree of similarity of random variables measured as a percentage of their common variance. Table \ref{tab4} shows the values of the determination coefficients given as a percentage. Their analysis shows that in most cases the analyzed variables are characterized by a low level of mutual similarity (mutual correlation). Only air temperature and dew point temperature are strongly correlated. In this case, the common variance measured by the coefficient of determination is over $ 76\% $. The coefficient of determination estimated for temperature and visibility indicates their common variance at a level slightly greater than $ 32\% $. The remaining determination coefficients do not exceed $ 10\% $.
	
	\begin{table}[t]
		\centering
		\caption{ Primary variables and their statistics }\label{tab2}
		\fontsize{8}{12}\selectfont{
			\begin{tabular}{ c||c|c|c|c|c|c|c }
				&$ X_1 $&$ X_2 $&$ X_3 $&$ X_4 $&$ X_5 $&$ X_6 $&$ X_7 $ \\ \cline{2-8}
				&Sea level&Air&Dew point &Wind&Wind&Visibility&Time of \\
				&pressure&temperature&temperature&direction&speed&&measurement \\ 
				&$ [mbar] $&$ [^{0}C] $&$ [^{0}C] $&$ [^0] $&$ [m/s] $&$ [m] $& \\ \hline \hline
				Mean&1016.378&10.221&5.314&180.805&3.397&18890.569&0.479 \\ \hline
				Median&1016.2&10&5.6&180&3&20000&0.46 \\ \hline
				Mode&1014.3&1.3&11.5&270&3&30000&0.75 \\ \hline
				Standard &\multirow{2}{*}{8.400}&\multirow{2}{*}{8.972}&\multirow{2}{*}{7.263}&\multirow{2}{*}{100.781}&\multirow{2}{*}{2.065}&\multirow{2}{*}{9769.928}&\multirow{2}{*}{0.289} \\
				deviation&&&&&&& \\ \hline
				Minimum&975.2&-18.6&-20.8&0&0&0&0 \\ \hline
				Maximum&1045.8&36.7&22.1&360&24&80000&0.96 \\ \hline
		\end{tabular}}
	\end{table}
	
	\begin{table}[t]
		\centering
		\caption{The matrix of correlation coefficients }\label{tab3}
		\fontsize{8}{12}\selectfont{
			\begin{tabular}{ c||c|c|c|c|c|c|c| }
				&$ x_1 $&$ x_2 $&$ x_3 $&$ x_4 $&$ x_5 $&$ x_6 $&$ x_7 $ \\ \hline \hline
				$ x_1 $&1&-0.197&-0.257&-0.110&-0.108&-0.032&-0.010 \\ \hline
				$ x_2 $&-0.197&1&0.875&0.025&-0.038&0.568&0.100 \\ \hline
				$ x_3 $&-0.257&0.875&1&0.031&-0.142&0.313&0.010 \\ \hline
				$ x_4 $&-0.110&0.025&0.031&1&0.311&0.050&0.034 \\ \hline
				$ x_5 $&-0.108&-0.038&-0.142&0.311&1&0.146&0.044 \\ \hline
				$ x_6 $&-0.032&0.568&0.313&0.050&0.146&1&0.122 \\ \hline
				$ x_7 $&-0.010&0.100&0.010&0.034&0.044&0.122&1 \\ \hline
			\end{tabular} \\[0.5cm]}
		\caption{The matrix of determination coefficients}\label{tab4}
		\fontsize{8}{12}\selectfont{
			\begin{tabular}{ c||c|c|c|c|c|c|c| }
				&$ x_1 $&$ x_2 $&$ x_3 $&$ x_4 $&$ x_5 $&$ x_6 $&$ x_7 $ \\ \hline \hline
				$ x_1 $&$100\%$&$3.89\%$&$6.63\%$&$1.21\%$&$1.17\%$&$0.10\%$&$0.01\%$\\ \hline
				$ x_2 $&$3.89\%$&$100\%$&$76.49\%$&$0.06\%$&$0.15\%$&$32.31\%$&$1.01\%$\\ \hline
				$ x_3 $&$6.63\%$&$76.49\%$&$100\%$&$0.09\%$&$2.03\%$&$9.77\%$&$0.01\%$\\ \hline
				$ x_4 $&$1.21\%$&$0.06\%$&$0.09\%$&$100\%$&$9.65\%$&$0.25\%$&$0.12\%$\\ \hline
				$ x_5 $&$1.17\%$&$0.15\%$&$2.03\%$&$9.65\%$&$100\%$&$2.12\%$&$0.20\%$\\ \hline
				$ x_6 $&$0.10\%$&$32.31\%$&$9.77\%$&$0.25\%$&$2.12\%$&$100\%$&$1.48\%$\\ \hline
				$ x_7 $&$0.01\%$&$1.01\%$&$0.01\%$&$0.12\%$&$0.20\%$&$1.48\%$&$100\%$\\ \hline
		\end{tabular}}
	\end{table}
	
	\begin{table}[t]
		\centering
		\caption{ Eigenvalues of the matrix of correlation coefficients }\label{tab5}
		\fontsize{8}{12}\selectfont{
			\begin{tabular}{ c||c|c|c|c|c|c|c| }
				Eigenvalue No.&1&2&3&4&5&6&7\\ \hline
				Eigenvalue&2.290&1.390&1.058&0.919&0.751&0.518&0.075\\ \hline
		\end{tabular}}
	\end{table}
	\begin{figure}[h!]
		\centering
		\includegraphics[width=0.65\textwidth]{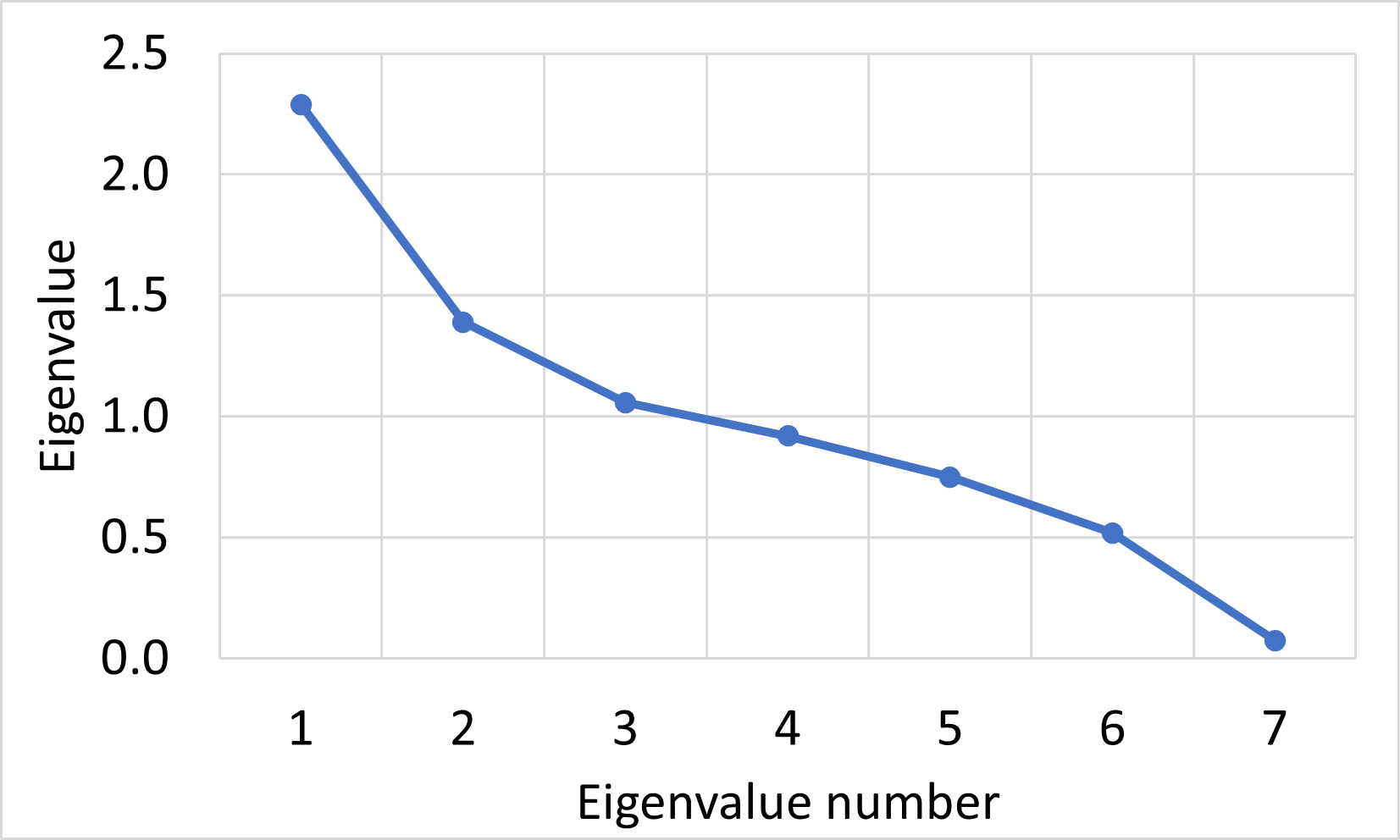}
		\caption{Scree plot for Dataset No. 1}\label{fig2}
	\end{figure}
	
	\subsection{Common elements in Principal Components Analysis and Factor Analysis}\label{commonElem}
	Both types of analyzes have common elements in their algorithms. The common element of both is solving the eigenproblem for the matrix of correlation coefficients. Therefore, this problem has been solved here. The non-increasing ordered eigenvalues obtained for the matrix of correlation coefficients from Table \ref{tab3} are presented in Table \ref{tab5}. Figure \ref{fig2} shows the scree plot for the obtained eigenvalues. The successive $ i- $th eigenvalue corresponds to the successive $ i- $th eigenvector $ U_i $:
	\begin{equation}\label{r45}
		U_i=\left[u_{1i},\cdots,u_{ni}\right]^T.
	\end{equation}
	Successive eigenvectors corresponding to the eigenvalues in Table \ref{tab5} form the columns of the matrix $ U $:
	\begin{equation}
		U=\left[U_1,\ldots,U_n\right]=\left[\begin{matrix}u_{11}&\cdots&u_{1n}\\\vdots&\ddots&\vdots\\u_{n1}&\cdots&u_{nn}\\\end{matrix}\right].
	\end{equation}
	For the considered data, the matrix of eigenvectors has the following form:
	\begin{equation}\label{r47}
		U=\left[\begin{matrix}-0.231&-0.218&0.579&0.579&-0.367&-0.306&0.030\\0.633&-0.097&0.028&0.084&-0.063&-0.193&-0.736\\0.583&-0.177&-0.187&-0.018&-0.219&-0.379&0.634\\0.067&0.625&-0.146&0.089&-0.716&0.251&-0.025\\0.005&0.696&0.057&0.197&0.432&-0.534&0.041\\0.438&0.122&0.383&0.350&0.319&0.609&0.228\\0.099&0.149&0.677&-0.699&-0.115&-0.081&0.042\\\end{matrix}\right].
	\end{equation}
	\subsection{Principal component analysis}
	The primary data is described in a standard coordinate system. Principal component analysis describes this data in a coordinate system defined by eigenvectors. Formula (\ref{r11}) finds the transition matrix ($ R $) from the standard coordinate system to the eigenvectors system.
	Formula (\ref{r15}) makes it possible to find the description of primary variables in the new coordinate system, thus finding a matrix containing the principal components $ P_C $:
	\begin{equation}\label{r49}
		P_C=x\cdot R^T.
	\end{equation}
	As a result of the transformation (\ref{r49}), seven principal components were obtained. These are uncorrelated random variables, the statistics of which are presented in Table \ref{tab5}. Comparing the results presented in Table \ref{tab5} with the results in Table \ref{tab4}, it can be seen that the variances of individual principal components are equal to the successive eigenvalues estimated for the correlation coefficient matrix contained in Table \ref{tab5}.
	\begin{table}[t]
		\centering
		\caption{Statistics of principal components}\label{tab6}
		\fontsize{8}{12}\selectfont{
			\begin{tabular}{ c||c|c|c|c|c|c|c| }
				&$ PC_1 $&$ PC_2 $&$ PC_3 $&$ PC_4 $&$ PC_5 $&$ PC_6 $&$ PC_7 $ \\ \hline \hline
				Mean&0.00&0.00&0.00&0.00&0.00&0.00&0.00\\ \hline
				Standard deviation&1.513&1.179&1.028&0.959&0.866&0.720&0.275\\ \hline
				Variance&2.290&1.390&1.058&0.919&0.751&0.518&0.075\\ \hline
		\end{tabular}}
	\end{table}
	
	\begin{table}[t]
		\centering
		\caption{ The correlation coefficients between the primary variables, and the principal components}\label{tab7}
		\fontsize{8}{12}\selectfont{
			\begin{tabular}{ c||c|c|c|c|c|c|c| }
				&$ PC_1 $&$ PC_2 $&$ PC_3 $&$ PC_4 $&$ PC_5 $&$ PC_6 $&$ PC_7 $ \\ \hline \hline
				$x_1$&-0.349&-0.257&0.595&0.555&-0.318&-0.220&0.008 \\ \hline
				$x_2$&0.957&-0.114&0.029&0.081&-0.054&-0.139&-0.202 \\ \hline
				$x_3$&0.882&-0.208&-0.193&-0.017&-0.190&-0.273&0.174 \\ \hline
				$x_4$&0.101&0.737&-0.150&0.085&-0.620&0.181&-0.007 \\ \hline
				$x_5$&0.008&0.820&0.058&0.189&0.375&-0.384&0.011 \\ \hline
				$x_6$&0.663&0.144&0.394&0.335&0.276&0.438&0.063 \\ \hline
				$x_7$&0.150&0.176&0.696&-0.670&-0.100&-0.058&0.012 \\ \hline 
			\end{tabular} \\[0.5cm]}
		\caption{ Percentage values of the coefficients of determination between primary variables and principal components}\label{tab8}
		\fontsize{8}{12}\selectfont{
			\begin{tabular}{ c||c|c|c|c|c|c|c| }
				&$ PC_1 $&$ PC_2 $&$ PC_3 $&$ PC_4 $&$ PC_5 $&$ PC_6 $&$ PC_7 $ \\ \hline \hline
				$x_1$&$12.21\%$&$6.60\%$&$35.43\%$&$30.82\%$&$10.09\%$&$4.84\%$&$0.01\%$ \\ \hline
				$x_2$&$91.65\%$&$1.31\%$&$0.08\%$&$0.65\%$&$0.29\%$&$1.93\%$&$4.08\%$\\  \hline
				$x_3$&$77.84\%$&$4.34\%$&$3.71\%$&$0.03\%$&$3.60\%$&$7.45\%$&$3.03\%$ \\ \hline
				$x_4$&$1.03\%$&$54.28\%$&$2.26\%$&$0.73\%$&$38.43\%$&$3.26\%$&$0.00\% $ \\ \hline
				$x_5$&$0.01\%$&$67.27\%$&$0.34\%$&$3.57\%$&$14.03\%$&$14.77\%$&$0.01\% $ \\ \hline
				$x_6$&$44.00\%$&$2.08\%$&$15.49\%$&$11.23\%$&$7.62\%$&$19.19\%$&$0.39\% $ \\ \hline
				$x_7$&$2.25\%$&$3.09\%$&$48.45\%$&$44.86\%$&$1.00\%$&$0.34\%$&$0.01\% $ \\ \hline
		\end{tabular}}
	\end{table}
	
	Having a set of primary variables and a set of principal components, the correlation coefficients between primary variables and principal components were estimated (Table \ref{tab7}). Based on the correlation coefficients, the coefficients of determination between the variables from both sets were found. Table \ref{tab8} contains information which variable and in what percentage is represented by successive principal components. It can be seen that most of the variances of the variables $ x_2 $ and $ x_3 $ represent the first principal component, $ PC_1 $.
	This component represents over $ 91\% $ of the variance of the $ x_2 $ variable and over $ 77\% $ of the variance of the $ x_3 $ variable. The second principal component $ PC_2 $ represents more than half of the variance of the variable $ x_4 $ and $ x_5 $. The common variance of these variables with the principal component $ PC_2 $ exceeds the level of $ 54\% $ and $ 67\% $, respectively. The variables $ x_1 $, $ x_6 $ and $ x_7 $ do not have a principal component that would represent most of their variance.
	For these variables, more components are needed to represent at least half of their variance.
	The principal components $ PC_3 $ and $ PC_4 $ represent most of the variances of the variables $ x_1 $ and $ x_7 $.
	In turn, the principal components $ PC_1 $ and $ PC_3 $ contain most of the variance of the variable $ x_6 $.
	
	Table \ref{tab9} also shows the coefficients of determination between primary variables and principal components, but now not in percent, but in absolute numbers. Additionally, it is enriched with sums of elements in rows and columns:
	\begin{itemize}
		\item The sum of the determination coefficients  in each row is equal to one. This is the variance of the standardized primary variable. The primary variable shared its variance with successive princupal components.
		\item The sum of the determination coefficients in each column is equal to the eigenvalue, i.e. the variance of the corresponding principal component. The principal component owes its variance to a certain part of the variance of the primary variables.
	\end{itemize}
	
	\begin{table}[t]
		\centering
		\caption{ Coefficients of determination between primary variables and principal components with sums in rows and columns}\label{tab9}
		\fontsize{8}{12}\selectfont{
			\begin{tabular}{ c||c|c|c|c|c|c|c||c|}
				&$ PC_1 $&$ PC_2 $&$ PC_3 $&$ PC_4 $&$ PC_5 $&$ PC_6 $&$ PC_7 $ &$\Sigma$ \\ \hline \hline
				$x_1$&0.122&0.066&0.354&0.308&0.101&0.048&0.000&1 \\ \hline
				$x_2$&0.917&0.013&0.001&0.007&0.003&0.019&0.041&1 \\ \hline
				$x_3$&0.778&0.043&0.037&0.000&0.036&0.075&0.030&1 \\ \hline
				$x_4$&0.010&0.543&0.023&0.007&0.384&0.033&0.000&1 \\ \hline
				$x_5$&0.000&0.673&0.003&0.036&0.140&0.148&0.000&1 \\ \hline
				$x_6$&0.440&0.021&0.155&0.112&0.076&0.192&0.004&1 \\ \hline
				$x_7$&0.022&0.031&0.484&0.449&0.010&0.003&0.000&1 \\ \hline \hline
				$\Sigma$&2.290&1.390&1.058&0.919&0.751&0.518&0.075&7.00\\  \hline
			\end{tabular} \\[0.5cm]}
		\caption{The cumulative variances of the primary variables represented by adding successive factors}\label{tab10}
		\fontsize{8}{12}\selectfont{
			\begin{tabular}{ c||c|c|c|c|c|c|c| }
				&$ PC_1 $&$ PC_2 $&$ PC_3 $&$ PC_4 $&$ PC_5 $&$ PC_6 $&$ PC_7 $ \\ \hline \hline
				$x_1$&$12.21\%$&$18.81\%$&$54.24\%$&$85.06\%$&$95.15\%$&$99.99\%$&$100\%$ \\ \hline
				$x_2$&$91.65\%$&$92.96\%$&$93.04\%$&$93.69\%$&$93.99\%$&$95.92\%$&$100\%$ \\ \hline
				$x_3$&$77.84\%$&$82.18\%$&$85.89\%$&$85.92\%$&$89.52\%$&$96.97\%$&$100\%$ \\ \hline
				$x_4$&$1.03\%$&$55.31\%$&$57.58\%$&$58.30\%$&$96.74\%$&$100.00\%$&$100\%$ \\ \hline
				$x_5$&$0.01\%$&$67.28\%$&$67.62\%$&$71.18\%$&$85.21\%$&$99.99\%$&$100\%$ \\ \hline
				$x_6$&$44.00\%$&$46.08\%$&$61.57\%$&$72.80\%$&$80.41\%$&$99.61\%$&$100\%$ \\ \hline
				$x_7$&$2.25\%$&$5.33\%$&$53.78\%$&$98.65\%$&$99.65\%$&$99.99\%$&$100\%$ \\ \hline
		\end{tabular}}
	\end{table}
	
	\begin{figure}[h!]
		\centering
		\includegraphics[width=0.65\textwidth]{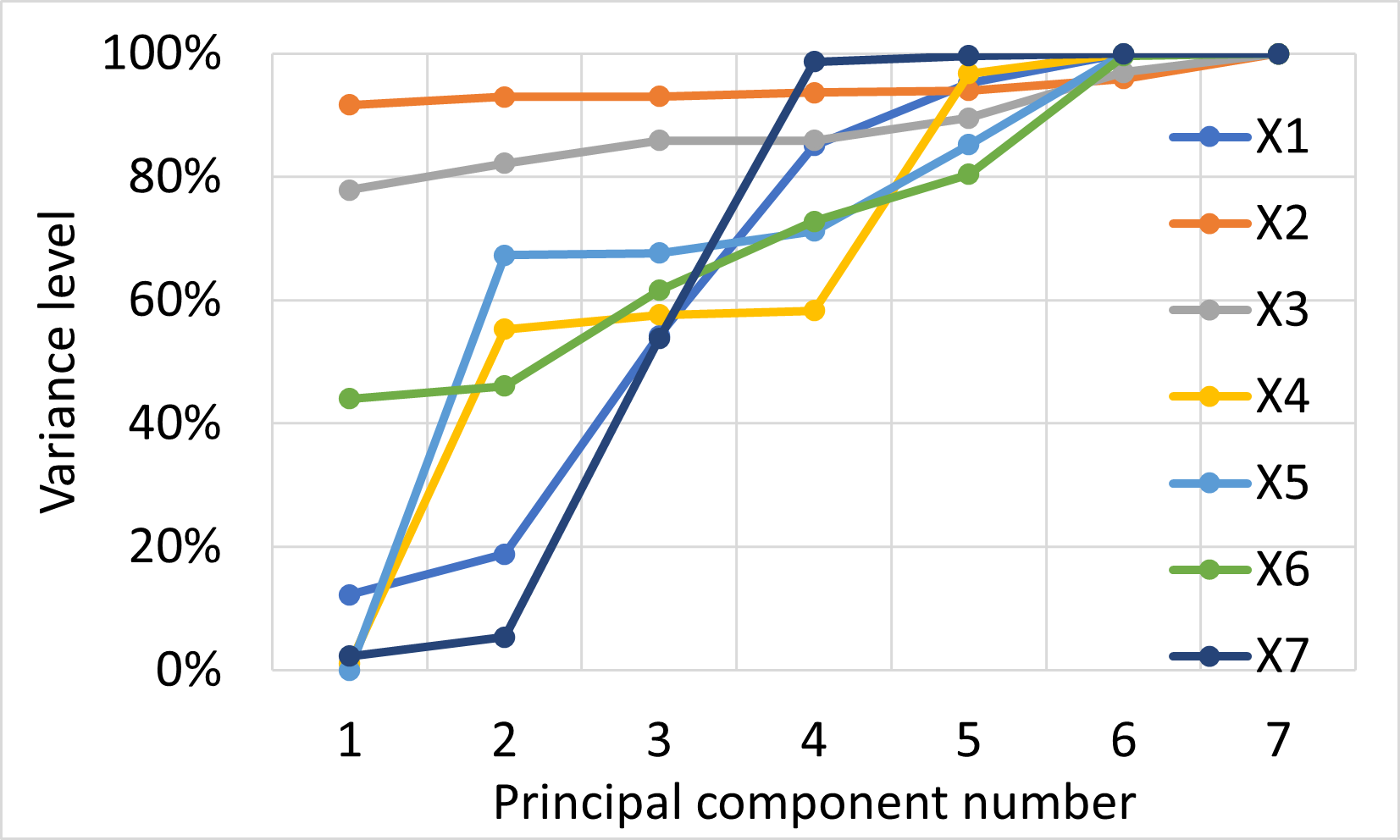}
		\caption{The level of representation of variances of primary variables by successive principal components}\label{fig3}
	\end{figure}
	
	Table \ref{tab10} presents the cumulative values of the coefficients of determination in the rows. 
	Cumulative values explain the level of representation of primary variables by successive principal components. 
	Figure \ref{fig3} shows a graphical representation of Table \ref{tab10}. 
	In Figure \ref{fig3} it is possible to see how many principal components are needed to achieve a satisfactory level of representation of the variance of individual primary variables. 
	It can be seen that one principal component $ PC_1 $ represents more than $ 70\% $ of the variance of the primary variable $ x_3 $ and more than $ 90\% $ of the variable $ x_2 $. 
	On the other hand, two principal components are not sufficient to represent more than half of the variance of the primary variables $ x_1 $ and $ x_7 $.
	\subsubsection{Geometric interpretation of principal components}\label{geomPCA}
	Correlation coefficients have the interpretation of cosines between the random components of the respective random variables \cite{Gniazdowski2013}. The square of the correlation coefficient between two random variables is called the coefficient of determination and measures the level of their common variance \cite{Gniazdowski2017}. In other words, the coefficient of determination measures the level of similarity of the random components of two random variables. On the other hand, if the random variables are independent, then the variance of the sum of these variables is equal to the sum of their variances \cite{Loeve1977}.
	
	Table \ref{tab7} contains the correlation coefficients between successive primary variables and principal components. Each of the rows in this table, as well as in Tables \ref{tab8} and \ref{tab9}, relates to a corresponding standardized primary variable. The coefficients of determination in each row of Table \ref{tab9} sum up to one, and thus to the variance of this standardized primary variable. This means that standardized primary variables can be decomposed into the sum of independent (orthogonal) random variables.
	
	Since the coefficient of determination measures the level of common variance of two random variables, the above-mentioned summed independent random variables are part of the principal components, and the variance of each (successive) of these independent random variables is the fraction of the variance of the successive principal components.
	
	The variance of a random variable is equal to the square of its standard deviation. Successive independent random variables can be interpreted geometrically as orthogonal anchored vectors at the origin of the coordinate system. The lengths of these vectors are equal to the standard deviations of these successive variables. The length squares of these vectors are equal to the variances of the random variables. Since vectors are orthogonal, on the basis of the Pythagorean theorem, the square of the length of their sum is the sum of the squares of their length. On the other hand, the length of the result vector is equal to the root of the sum of the squares of the lengths of the summed vectors.
	
	The above analysis leads to the geometric interpretation of PCA in the vector space \cite{Gniazdowski2017}. The correlation coefficients between the $ i-th $ primary variable and the following principal components ($ i-th $ row in Table \ref{tab7}) can be interpreted as components of the vector representing the primary variables in the coordinate system composed of eigenvectors. Due to the existing analogy between independent random variables and orthogonal vectors, a geometric interpretation can be used to describe the behavior of such random variables, and in particular, the Pythagorean theorem can be used\footnote{Analogous vector interpretation can be applied to principal components. Each $ j-th $ principal component can be decomposed into the sum of mutually independent (orthogonal) random variables. In the same way, the variance of each principal component consists of the variance of these mutually independent (orthogonal) components. The correlation coefficients between the $ j-th $ principal component and successive primary variables ($ j-th $ column in Table \ref{tab7}) can be interpreted as components of the vector representing the principal components in the standard coordinate system.}. Since each row is a vector representing a single primary variable, the cosines between successive vectors are identical to the correlation coefficients between the successive primary variables that these vectors represent.
	\subsubsection{Determination of the number of principal components }\label{selectInPCA}
	In PCA, there are as many components as there are primary variables. However, not all of them should be identified. Principal components that carry a minimal amount of information can be omitted. This means that those with the smallest variance can be ignored. The rejection of the principal components with the smallest variance leads to a reduction of the space dimension in the case of the above-described vector representation of primary variables by principal components. Leaving the $ k $ principal components with the largest variance leads to the rejection of the $ n-k $ principal components with the smallest variance. The rejection of the $ n-k $ principal components is equivalent to the rejection of the $ n-k $ columns from Table \ref{tab7}.
	
	In order to solve the problem of determining the appropriate number of principal components, various previously known criteria for determining their number should be discussed:
	\begin{itemize}
		\item The scree plot criterion does not apply here as the scree plot (Figure \ref{fig1}) does not show two phases that are clearly separated by a so-called ''elbow''. The first phase of rapid descent and the second phase of gentle descent are missing here.
		\item 	The percentage criterion of the part of the variance explained by the principal components requires examination of Table \ref{tab11}. This table contains the distribution of variance explained by the successive principal components.  It is assumed that there should be so many principal components that the sum of the eigenvalues associated with successive principal components is not less than the specified percentage threshold in relation to the trace of the correlation matrix. The selection of the three principal components will explain less than $ 70\% $ of the variance of the primary variables.  If we assume that the principal components should explain at least $ 80\% $ of the variance of the primary variables, then the four principal components satisfactorily meet this criterion.
		\item  	Since the third eigenvalue is the smallest eigenvalue, not less than one, the Kaiser criterion suggests a selection of three principal components.
		\item The criterion of half the number of primary variables suggests choosing three out of seven principal components.
	\end{itemize}
	As can be seen, different criteria used can lead to different results. One more criterion will therefore be examined here. This criterion, called the ''Minimum Communality Criterion'', was signaled enigmatically in the book \cite{Larose2006}. A similar criterion was independently proposed in \cite{Gniazdowski2017}. This criterion will be shown in the example presented in Table \ref{tab12}. It was assumed that the primary variables $ x_1, \ldots, x_7 $ will be represented by the three principal components $ PC_1 $, $ PC_2 $ and $ PC_3 $.
	The table shows what level of variance of the $ x_i $ variable is represented by the prinipal component $ PC_j $. For example, $ 91.65\% $ of the variance of the primary variable $ x_2 $ is represented by the principal component $ PC_1 $. The same principal component represents the primary variables $ x_4 $, $ x_5 $ and $ x_7 $ to a very small extent. The level of representation of these variables by the principal component $ PC_1 $ amounts to $ 1.03\% $, $ 0.01\% $ and $  2.27\% $, respectively.
	The row marked as ''Average in column'' in Table \ref{tab12} is identical to the column ''Percentage of variance explained by each PC'' in Table \ref{tab10}. Table \ref{tab11} refers to the eigenvalues. Eigenvalues are identical to variances.
	A comparison of Table \ref{tab11} and Table \ref{tab12} shows that the results in Table \ref{tab11}  refer to the mean variance of the primary variables explained by each principal component. Looking at the last row in the last column of Table \ref{tab12}, it can be seen that the three principal components contain just over $ 67\% $ of the mean variance of all primary variables. However, the variance of single primary variables is represented to a varying degree. The variance of the primary variables $ x_1 $, $ x_4 $ and $ x_7 $ is represented in slightly more than half, and the variance of the primary variable $ x_2 $ is represented in more than $ 93\% $.
	
	\begin{table}[t]
		\centering
		\caption{The percentage of variances explained by the successive principal components}\label{tab11}
		\fontsize{8}{12}\selectfont{
			\begin{tabular}{c|c|c|c|c} \hline 
				\multirow{2}{*}{No.} & \multirow{2}{*}{Eigenvalue} & Cumulative & Percentage of variance & Cumulative \\ 
				& & eigenvalues & explained by each PC & percentage of variance \\ \hline \hline
				1&$2.290$&$2.290$&$32.71\%$&$32.71\%$ \\ \hline
				2&$1.390$&$3.680$&$19.85\%$&$52.56\%$ \\ \hline
				3&$1.058$&$4.737$&$15.11\%$&$67.67\%$ \\ \hline
				4&$0.919$&$5.656$&$13.13\%$&$80.80\%$ \\ \hline
				5&$0.751$&$6.407$&$10.72\%$&$91.52\%$ \\ \hline
				6&$0.518$&$6.925$&$7.40\%$&$98.92\%$ \\ \hline
				7&$0.075$&$7.000$&$1.08\%$&$100.00\%$ \\ \hline	
		\end{tabular}}
	\end{table}
	
	\begin{table}[t]
		\centering
		\caption{ The level of representation of primary variables by the three principal components}\label{tab12}
		\fontsize{8}{12}\selectfont{
			\begin{tabular}{ c||c|c|c||c| }
				&$ PC_1 $&$ PC_2 $&$ PC_3 $&$\Sigma$ \\ \hline \hline
				$x_1$&$12.21\%$&$6.60\%$&$35.43\%$&$54.24\%$ \\ \hline
				$x_2$&$91.65\%$&$1.31\%$&$0.08\%$&$93.04\%$ \\ \hline
				$x_3$&$77.84\%$&$4.34\%$&$3.71\%$&$85.89\%$ \\ \hline
				$x_4$&$1.03\%$&$54.28\%$&$2.26\%$&$57.58\%$ \\ \hline
				$x_5$&$0.01\%$&$67.27\%$&$0.34\%$&$67.62\%$ \\ \hline
				$x_6$&$44.00\%$&$2.08\%$&$15.49\%$&$61.57\%$ \\ \hline
				$x_7$&$2.25\%$&$3.09\%$&$48.45\%$&$53.78\%$ \\ \hline \hline
				Average in column&$32.71\%$&$19.85\%$&$15.11\%$&$67.67\%$ \\ \hline
		\end{tabular}}
	\end{table}
	
	The above observations confirm that there is an additional criterion for determining the appropriate number of principal components with regard to the degree of reconstruction of the variance of the primary variables. The application of this criterion will allow to determine the appropriate number of principal components in such a way that the level of variance representation of each of the primary variables is at least satisfactory, not lower than the set threshold \cite{Gniazdowski2017}. There should be enough principal components so that most of the variance of each of the primary variables can be reproduced. Common sense suggests that most means more than half the variance.
	
	Of course, there still remains the technical problem of applying this criterion. The criterion presented here requires identifying all the principal components, then calculating the correlation coefficients and the coefficients of determination between the primary variables and the principal components, then determining which principal components are necessary and rejecting the others. Compared to the previously known criteria, the criterion presented here has greater computational complexity, both in terms of time and memory complexity:
	\begin{itemize}
		\item The proposed criterion in the form presented above has a greater time complexity than the previous criteria, because it requires the identification of all principal components (not only selected ones), and then requires the estimation of correlation coefficients between all standardized primary variables and all principal components. For $ n $ primary variables and $ n $ principal components, it is also necessary to estimate $ n^2 $ correlation coefficients.
		\item 	The greater complexity of memory manifests itself in the fact that before identifying the final set of principal components, all principal components must first be identified.
	\end{itemize}
	The criteria in subsection \ref{selectCommon} are not that complex. They allow you to make a decision regarding the selection of principal components.
	After making a decision, it is enough to identify only selected principal components. For this purpose, before performing the operation (\ref{r49}), it is enough to discard as many last columns from the matrix $ R^T $ as should be discarded of principal components. On the other hand, the disadvantage of these criteria is that they do not always allow the identification of as many principal components as to be able to present most of the variance of each of the primary variables.
	
	Due to the reduction in the number of major components, attention should be paid to the consequences of this reduction. Since fundamental variables are described as vectors, reducing the number of principal components is equivalent to reducing the size of the space in which the vectors are described. Reducing the size of the space can simplify the analysis that is performed.
	
	\begin{table}[t]
		\centering
		\caption{ Full matrix of factor loadings}\label{tab13}
		\fontsize{8}{12}\selectfont{
			\begin{tabular}{ c||c|c|c|c|c|c|c| }
				&$ F_1 $&$ F_2 $&$ F_3 $&$ F_4 $&$ F_5 $&$ F_6 $&$ F_7 $ \\ \hline \hline
				$x_1$&-0.349&-0.257&0.595&0.555&-0.318&-0.220&0.008 \\ \hline
				$x_2$&0.957&-0.114&0.029&0.081&-0.054&-0.139&-0.202 \\ \hline
				$x_3$&0.882&-0.208&-0.193&-0.017&-0.190&-0.273&0.174 \\ \hline
				$x_4$&0.101&0.737&-0.150&0.085&-0.620&0.181&-0.007 \\ \hline
				$x_5$&0.008&0.820&0.058&0.189&0.375&-0.384&0.011 \\ \hline
				$x_6$&0.663&0.144&0.394&0.335&0.276&0.438&0.063 \\ \hline
				$x_7$&0.150&0.176&0.696&-0.670&-0.100&-0.058&0.012 \\ \hline
		\end{tabular}}
	\end{table}
	\begin{table}[h!]
		\centering
		\caption{Cumulative matrix of common variances }\label{tab14}
		\fontsize{8}{12}\selectfont{
			\begin{tabular}{ c||c|c|c|c|c|c|c| }
				&$ F_1 $&$ F_2 $&$ F_3 $&$ F_4 $&$ F_5 $&$ F_6 $&$ F_7 $ \\ \hline \hline
				$x_1$&0.1221&0.1881&0.5424&0.8506&0.9515&0.9999&1 \\ \hline
				$x_2$&0.9165&0.9296&0.9304&0.9369&0.9399&0.9592&1 \\ \hline
				$x_3$&0.7784&0.8218&0.8589&0.8592&0.8952&0.9697&1 \\ \hline
				$x_4$&0.0103&0.5531&0.5758&0.5830&0.9674&1.0000&1 \\ \hline
				$x_5$&0.0001&0.6728&0.6762&0.7118&0.8521&0.9999&1 \\ \hline
				$x_6$&0.4400&0.4608&0.6157&0.7280&0.8041&0.9961&1 \\ \hline
				$x_7$&0.0225&0.0533&0.5378&0.9865&0.9965&0.9999&1 \\ \hline
		\end{tabular}}
	\end{table}
	
	\subsection{Factor analysis}\label{vectFA}
	Factor analysis was also performed for Dataset No. 1. Using the eigenvalues estimated for the matrix of correlation coefficients (Table \ref{tab5}) and the eigenvectors (\ref{r47}) corresponding to these eigenvalues, the full matrix of factor loadings (\ref{r27}) was found using the formulas (\ref{r25}) and (\ref{r26}). This matrix is presented in Table \ref{tab13}. It can be seen that the content of this table is identical to the content of Table \ref{tab7}. Table \ref{tab7} contains the correlation coefficients between the primary variables and the principal components obtained in the PCA. Table \ref{tab13} contains the complete matrix of factor loadings obtained in the FA. The identity of tables \ref{tab7} and \ref{tab13} means that each factor loading that connects the $ i-t $h primary variable to the $ j-th $ factor is equal to the correlation coefficient between the $ i-th $ primary variable and the $ j-th $ principal component. It means that:
	\begin{itemize}
		\item The factors obtained in the factor analysis can be identified before their rotation with the standardized principal components obtained in the principal components analysis.
		\item 	Factors connecting the i-th primary variable with successive principal components (i-th row in Table \ref{tab13}) can be interpreted as components of the vector representing the primary variables in the coordinate system made up of eigenvectors. Thanks to this, a geometric description can be used to describe the behavior of primary variables, in particular, the Pythagorean theorem can be used.
		\item Since in the vector interpretation each single row is a vector representing a single primary variable, therefore, as in PCA, the cosines between successive vectors are identical to the correlation coefficients between those primary variables that these vectors represent.
	\end{itemize}
	\subsubsection{Artifact}\label{artefakt}
	During the analysis of the full matrix of factor loadings $ L $ (\ref{r27}), a fact was observed, which will be presented here in more detail\footnote{By the way, it should be mentioned that the article \cite{Gniazdowski2017} describes an analogous fact that was observed in the context of the analysis of the matrix containing the correlation coefficients between the primary variables and the principal components.}. For this purpose, attention should be paid to some properties of this full matrix of factor loadings:
	\begin{itemize}
		\item The rows of the full matrix of factor loadings $ L $ can be interpreted as vectors that represent successive standardized primary variables. The sums of the squares of the components of the row vectors, representing the squares of the lengths of these vectors, are equal to the unit variances of the standardized primary variables.
		\item The columns of the $ L $ matrix can be interpreted as vectors that represent the principal components in PCA. The sums of the squares of the components of the column vectors representing the squares of the lengths of these vectors are equal to the eigenvalues of the correlation coefficient matrix, and thus equal to the variances of the principal components in the PCA.
	\end{itemize}
	By multiplying the factor loadings matrix  $ L $ by the transposition of the eigenvector matrix $ U^T $, the following symmetric matrix was obtained:
	\begin{equation}\label{r14a}
		L\cdot U^T=\left[\begin{matrix}
			0.987&-0.076&-0.122&-0.050&-0.057&0.004&-0.003\\
			-0.076&0.803&0.508&0.005&-0.014&0.297&0.050\\
			-0.122&0.508&0.843&0.018&-0.084&0.095&-0.011\\
			-0.050&0.005&0.018&0.986&0.157&0.018&0.015\\
			-0.057&-0.014&-0.084&0.157&0.979&0.080&0.019\\
			0.004&0.297&0.095&0.018&0.080&0.945&0.055\\
			-0.003&0.050&-0.011&0.015&0.019&0.055&0.997\\
		\end{matrix}\right].
	\end{equation}
	The $ U^T $ matrix, similarly to the $  U $ (\ref{r25}) matrix, is an orthogonal matrix, and therefore describes a certain rotation of the coordinate system in which the row vectors of the factor loadings matrix are described.
	Rotation means changing the basis, or in other words changing the coordinate system in which the vectors are described.
	So there is a new basis in which the factor loadings matrix is symmetrical.
	This means that not only are the sums of the squares of the row vector components equal to $ 1 $, but also the sums of the squares of the column vector components are equal to $ 1 $.
	As a result of the performed rotation, the row vectors representing standardized primary variables did not change.
	It only happened that the standardized primary variables are represented by a different set of factors than before the rotation.
	After rotation, the primary variables can be described as linear combinations of independent factors, but not factors identical to the standardized principal components.
	Now the factors are random variables with unit variances.
	
	With regard to the artifact described here, questions arise about both its causes and its potential effects. As for the effects, it is still unknown whether they are important from the point of view of data analysis. As for the causes, the article \cite{Gniazdowski2017} did not know them yet. In the context of FA, it seems that more can now be said about the causes. An attempt to explain the causes will be undertaken in section 5 where some detailed results presented in this paper will be discussed.
	\subsubsection{Determining the number of factors}\label{selectInFA}
	Section \ref{selectCommon} presents the basic methods for determining the number of principal components in PCA, as well as methods for determining the number of factors in factor analysis.
	In subsection \ref{selectInFA}, the criteria from subsection \ref{selectCommon} were used to select the principal components, pointing to differences in the results of their operation.
	In FA for the same data, the above criteria will produce the same results as for PCA.
	This is due to the fact that the criteria discussed here both in PCA and in FA refer to the eigenproblem solved for the same matrix of correlation coefficients, i.e. to identical eigenvalues.
	
	On the other hand, section \ref{selectInFA}  also suggests a new criterion for determining the number of principal components. It was noted that this criterion has a much greater computational complexity than each of the criteria from subsection \ref{selectCommon}. The proposed criterion first requires the identification of all principal components, and then it requires the estimation of the correlation coefficients between all principal components and all primary variables.
	
	In the case of FA, it can be hoped that this new criterion will be characterized by a lower computational complexity.
	Although Table \ref{tab13} representing the full matrix of factor loadings is identical to Table \ref{tab7}, which represents the matrix of correlation coefficients between primary variables and principal components, in the case of FA, the method of its estimation does not require calculating the correlation coefficients, but only performing the operation (\ref{r27}), i.e. scaling successive eigenvectors by roots of successive eigenvalues.
	
	The squares of the factor loadings are a measure of the common variance between the standardized primary variable and the factor. As the factors are independent, the common variances can be summed up in each row. For each row, Table \ref{tab14} shows the cumulative common variances from successive factors. These cumulative variances explain the level of representation of each standardized primary variable by successive factors. It can be seen that in each row, the sum of successive variances tends to one, that is, to the variance of the standardized primary variable.
	
	Table \ref{tab15} is a copy of Table \ref{tab14}. The difference is that in Table \ref{tab15} the cell content is shown as a percentage.
	The last row has also been added to Table \ref{tab15}, which contains the mean values calculated for each successive column. The content of this row is identical to the content of the last column in table \ref{tab11}, named ''Cumulative percentage of variance''. 
	The analysis of the last column in table 11 in the context of the last row of table \ref{tab15} allows to evaluate the level of reconstruction of the mean variance of all standardized primary variables by successive factors. Thus, one factor explains $ 32.71\% $ of the mean variance of all primary variables. When analyzing Table \ref{tab15}, it can also be seen that the first factor explains $ 95.65\% $ of the variance of the primary variable $ x_2 $, but the same factor only explains about $ 0.01\% $ of the variance of the primary variable $ x_5 $. Two factors explain $ 52.56\% $ of the mean variance of all primary variables. The same two factors explain $ 92.96\% $ of the variance of the primary variable $ x_2 $ and only $ 5.33\% $ of the variance of the primary variable $ x_7 $. In the same way, the influence of the sucessive factors on the reconstruction of the mean variance of all primary variables as well as on the reconstruction of the variance of individual primary variables can be analyzed.
	
	\begin{table}[t]
		\centering
		\caption{ Cumulative matrix of common variances as percentages, considering the mean value for each column}\label{tab15}
		\fontsize{8}{12}\selectfont{
			\begin{tabular}{ c||c|c|c|c|c|c|c| }
				&$ F_1 $&$ F_2 $&$ F_3 $&$ F_4 $&$ F_5 $&$ F_6 $&$ F_7 $ \\ \hline \hline
				$x_1$&$12.21\%$&$18.81\%$&$54.24\%$&$85.06\%$&$95.15\%$&$99.99\%$&$100\%$ \\ \hline
				$x_2$&$91.65\%$&$92.96\%$&$93.04\%$&$93.69\%$&$93.99\%$&$95.92\%$&$100\%$ \\ \hline
				$x_3$&$77.84\%$&$82.18\%$&$85.89\%$&$85.92\%$&$89.52\%$&$96.97\%$&$100\%$ \\ \hline
				$x_4$&$1.03\%$&$55.31\%$&$57.58\%$&$58.30\%$&$96.74\%$&$100.00\%$&$100\%$ \\ \hline
				$x_5$&$0.01\%$&$67.28\%$&$67.62\%$&$71.18\%$&$85.21\%$&$99.99\%$&$100\%$ \\ \hline
				$x_6$&$44.00\%$&$46.08\%$&$61.57\%$&$72.80\%$&$80.41\%$&$99.61\%$&$100\%$ \\ \hline
				$x_7$&$2.25\%$&$5.33\%$&$53.78\%$&$98.65\%$&$99.65\%$&$99.99\%$&$100\%$ \\ \hline \hline
				Average in column&$32.71\%$&$52.56\%$&$67.67\%$&$80.80\%$&$91.52\%$&$98.92\%$&$100\%$ \\ \hline
		\end{tabular}}
	\end{table}
	\begin{table}
		\centering
		\caption{Minimum variance (MinVar) and mean variance (AverVar) reproduced by successive factors}\label{tab16}
		\fontsize{8}{12}\selectfont{
			\begin{tabular}{ c||c|c|c|c|c|c|c| }
				No. of factors&1&2&3&4&5&6&7 \\ \hline 
				EigVal&$32.71\%$&$19.85\%$&$15.11\%$&$13.13\%$&$10.72\%$&$7.40\%$&$1.08\%$ \\ \hline
				MinVar&$0.01\%$&$5.33\%$&$53.78\%$&$58.30\%$&$80.41\%$&$95.92\%$&$100\%$ \\ \hline
				AverVar&$32.71\%$&$52.56\%$&$67.67\%$&$80.80\%$&$91.52\%$&$98.92\%$&$100\%$ \\ \hline
				NrMinVar&5&7&7&4&6&2&6 \\ \hline
		\end{tabular}}
	\end{table}
	\begin{figure}[th!]
		\centering
		\includegraphics[width=0.65\textwidth]{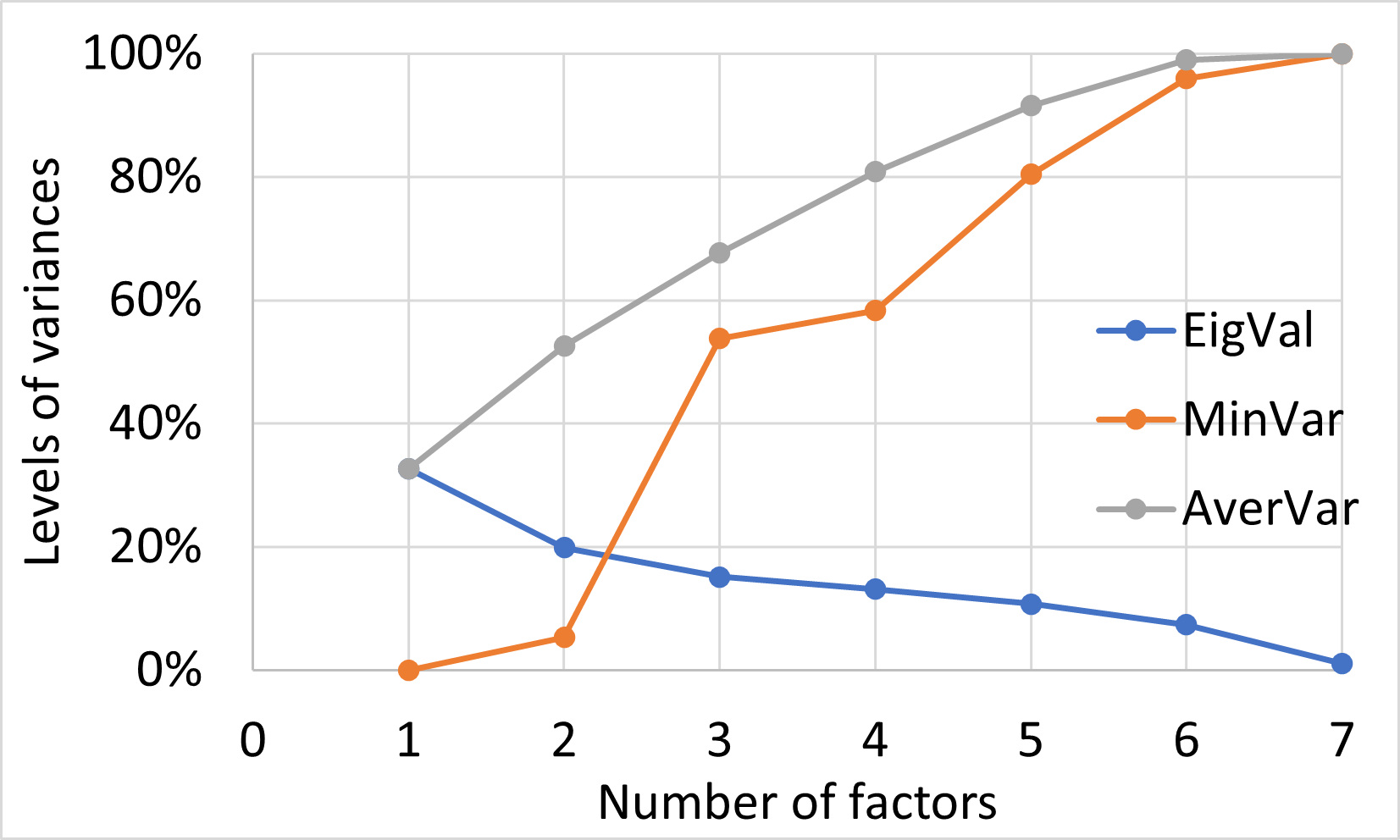}
		\caption{Minimum variance (MinVar) and mean variance (AverVar) reproduced by successive factors}\label{fig4}
	\end{figure}
	
	Based on the content of Table \ref{tab15}, Table \ref{tab16} was created, in which for successive factors, the following lines present:
	\begin{itemize}
		\item EigVal -- successive eigenvalues given as a percentage of the trace of the matrix of correlation coefficients,
		\item MinVar -- the level of reconstruction of the variance of the variable least represented by successive factors,
		\item AverVar -- the average level of variance of all primary variables explained by successive factors,
		\item NrMinVar -- number of the variable whose variance is least represented by successive factors.
	\end{itemize}
	The content of Table \ref{tab16} (excluding the last line) is shown in Figure \ref{fig4}. It can be seen that the line labeled ''EigVal'' is the scaled scree plot shown in Figure \ref{fig2}.
	
	For any single primary variable, it is wise to represent most of its variance. Most of course means more than half the variance. Assuming the minimum level of reproduction of individual primary variables, it is possible to read from the chart or table how many factors will meet the assumed level of reproduction of the variance of individual primary variables. In this case, the 3 factors will allow to reproduce $ 53.78\% $ of the variance of the primary variable $ x_7 $.
	
	The presented analysis showed that, depending on the adopted criteria, the factor model should contain three or four factors. Three factors result from the Kaiser criterion and the criterion of half the number of primary variables. On the other hand, these three factors are the minimum number of factors that can reproduce most of the variance of each of the primary variables.
	
	Finally, it can be seen that reducing the number of factors has some consequences.
	Since the primary variables are interpreted as vectors, any reduction in the number of factors is equivalent to a reduction in the size of the space in which the factor analysis is performed, which may simplify the factor analysis.
	
	\begin{table}[t]
		\centering
		\caption{The matrix of factor loadings for a three-factor model}\label{tab17}
		\fontsize{8}{12}\selectfont{
			\begin{tabular}{ c||c|c|c||c| }
				&$ F_1 $&$ F_2 $&$ F_3 $&Communality\\ \hline \hline
				$x_1$&-0.3494&-0.2569&0.5952&$54.24\%$\\  \hline
				$x_2$&0.9574&-0.1143&0.0286&$93.04\%$\\  \hline
				$x_3$&0.8823&-0.2083&-0.1927&$85.89\%$\\  \hline
				$x_4$&0.1015&0.7368&-0.1504&$57.58\%$\\  \hline
				$x_5$&0.0082&0.8202&0.0583&$67.62\%$\\  \hline
				$x_6$&0.6634&0.1442&0.3935&$61.57\%$\\  \hline
				$x_7$&0.1499&0.1757&0.6960&$53.78\%$\\   \hline 
			\end{tabular} \\[0.5cm]}
		\caption{The matrix of common variances for a three-factor model}\label{tab18}
		\fontsize{8}{12}\selectfont{
			\begin{tabular}{c||c|c|c||c|}
				&$ F_1 $&$ F_2 $&$ F_3 $&Communality\\ \hline \hline
				$x_1$&$12.21\%$&$6.60\%$&$35.43\%$&$54.24\%$\\  \hline
				$x_2$&$91.65\%$&$1.31\%$&$0.08\%$&$93.04\%$\\  \hline
				$x_3$&$77.84\%$&$4.34\%$&$3.71\%$&$85.89\%$\\  \hline
				$x_4$&$1.03\%$&$54.28\%$&$2.26\%$&$57.58\%$\\  \hline
				$x_5$&$0.01\%$&$67.27\%$&$0.34\%$&$67.62\%$\\  \hline
				$x_6$&$44.00\%$&$2.08\%$&$15.49\%$&$61.57\%$\\  \hline
				$x_7$&$2.25\%$&$3.09\%$&$48.45\%$&$53.78\%$\\  \hline
		\end{tabular}}
	\end{table}
	
	\subsubsection{Rotation of the model with three factors}
	Assuming that the primary variables can be modeled with three factors, Table \ref{tab17} shows the factor loadings matrix for the three-factor model. 
	Table \ref{tab18} shows the common variances between the primary variables and the three factors. 
	The table shows that the primary variables $ x_2 $ and $ x_3$ are significantly similar to the first factor. 
	On the other hand, most of the variances of the primary variables $ x_4 $ and $ x_5 $ are represented by the second factor. 
	Unfortunately, it is impossible to indicate which factor carries most of the variances of the primary variables $ x_1 $, $ x_6 $ and $ x_7 $.
	
	After the Varimax rotation for the three-factor model (Table 19), only a slight improvement was obtained. When analyzing the table of common variances after obtained after the Varimax rotation (Table 20), it can be seen that the third factor represents the majority of the variance of the $ x_7 $ variable. Unfortunately, the variables $ x_1 $ and $ x_6 $ still do not have the dominant factor that would represent most of their variance.
	
	\begin{table}[t]
		\centering
		\caption{The matrix of factor loadings for a three-factor model after Varimax rotation}\label{tab19}
		\fontsize{8}{12}\selectfont{
			\begin{tabular}{ c||c|c|c||c| }
				&$ F_1 $&$ F_2 $&$ F_3 $&Communality\\ \hline \hline
				$x_1$&-0.3314&-0.3410&0.5624&$54.24\%$\\  \hline
				$x_2$&0.9634&-0.0250&0.0403&$93.04\%$ \\ \hline
				$x_3$&0.9009&-0.1056&-0.1901&$85.89\%$\\  \hline
				$x_4$&0.0320&0.7536&-0.0831&$57.58\%$\\  \hline
				$x_5$&-0.0719&0.8088&0.1300&$67.62\%$\\  \hline
				$x_6$&0.6406&0.1708&0.4196&$61.57\%$ \\ \hline
				$x_7$&0.1223&0.1263&0.7120&$53.78\%$\\  \hline
			\end{tabular} \\[0.5cm]}
		\caption{The matrix of common variances for a three-factor model after Varimax rotation}\label{tab20}
		\fontsize{8}{12}\selectfont{
			\begin{tabular}{ c||c|c|c||c| }
				&$ F_1 $&$ F_2 $&$ F_3 $&Communality\\ \hline \hline
				$x_1$&$10.98\%$&$11.63\%$&$31.63\%$&$54.24\%$\\ \hline
				$x_2$&$92.82\%$&$0.06\%$&$0.16\%$&$93.04\%$ \\ \hline
				$x_3$&$81.16\%$&$1.12\%$&$3.61\%$&$85.89\%$\\ \hline
				$x_4$&$0.10\%$&$56.78\%$&$0.69\%$&$57.58\%$\\ \hline
				$x_5$&$0.52\%$&$65.41\%$&$1.69\%$&$67.62\%$\\ \hline
				$x_6$&$41.04\%$&$2.92\%$&$17.61\%$&$61.57\%$\\ \hline
				$x_7$&$1.50\%$&$1.60\%$&$50.69\%$&$53.78\%$\\ \hline
		\end{tabular}}
	\end{table}

	\begin{figure}[t]
		\centering
		\includegraphics[width=0.65\textwidth]{ 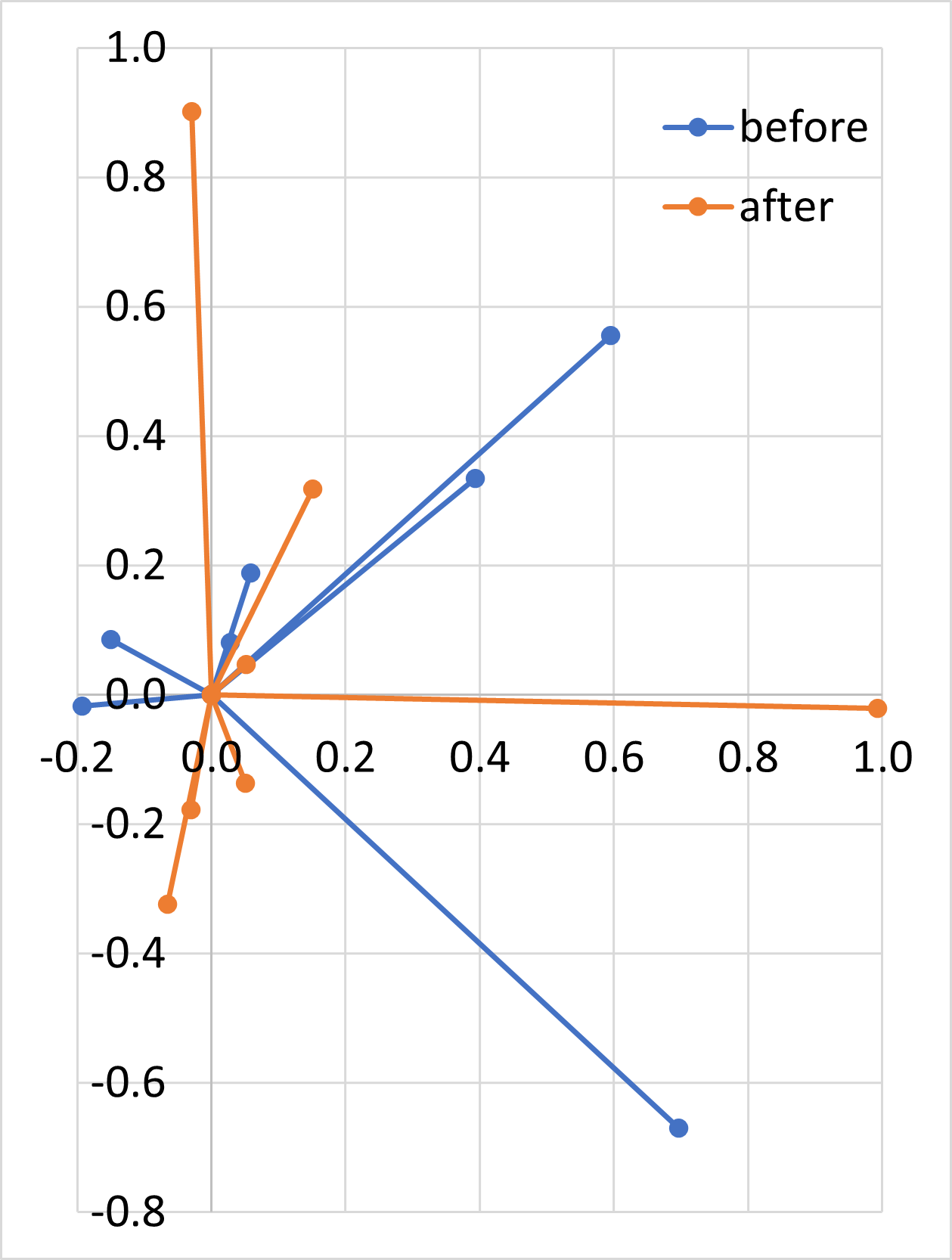}
		\caption{ Rotation on the $ x_3\times x_4 $ plane for the four factor model}\label{fig5}
	\end{figure}
	
	\begin{table}[t]
		\centering
		\caption{The matrix of factor loadings for a four-factor model}\label{tab21}
		\fontsize{8}{12}\selectfont{
			\begin{tabular}{ c||c|c|c|c||c| }
				&$ F_1 $&$ F_2 $&$ F_3 $&$ F_4 $&Communality\\ \hline \hline
				$x_1$&-0.349&-0.257&0.595&0.555&0.851\\ \hline
				$x_2$&0.957&-0.114&0.029&0.081&0.937\\ \hline
				$x_3$&0.882&-0.208&-0.193&-0.017&0.859\\ \hline
				$x_4$&0.101&0.737&-0.150&0.085&0.583\\ \hline
				$x_5$&0.008&0.820&0.058&0.189&0.712\\ \hline
				$x_6$&0.663&0.144&0.394&0.335&0.728\\ \hline
				$x_7$&0.150&0.176&0.696&-0.670&0.986\\ \hline
			\end{tabular} \\[0.5cm]}
		
		\caption{The matrix of common variances for a four-factor model}\label{tab22}
		\fontsize{8}{12}\selectfont{
			\begin{tabular}{ c||c|c|c|c||c| }
				&$ F_1 $&$ F_2 $&$ F_3 $&$ F_4 $&Communality\\ \hline \hline
				$x_1$&$12.21\%$&$6.60\%$&$35.43\%$&$30.82\%$&$85.06\%$\\ \hline
				$x_2$&$91.65\%$&$1.31\%$&$0.08\%$&$0.65\%$&$93.69\%$\\ \hline
				$x_3$&$77.84\%$&$4.34\%$&$3.71\%$&$0.03\%$&$85.92\%$\\ \hline
				$x_4$&$1.03\%$&$54.28\%$&$2.26\%$&$0.73\%$&$58.30\%$\\ \hline
				$x_5$&$0.01\%$&$67.27\%$&$0.34\%$&$3.57\%$&$71.18\%$\\ \hline
				$x_6$&$44.00\%$&$2.08\%$&$15.49\%$&$11.23\%$&$72.80\%$\\ \hline
				$x_7$&$2.25\%$&$3.09\%$&$48.45\%$&$44.86\%$&$98.65\%$\\ \hline
		\end{tabular}}
	\end{table}

	\subsubsection{Rotation of the factor model with the four factors}
	As in the model with three factors, even after the rotation, it was not possible to reach a situation where most of the variance of each of the primary variables would be represented by one of the factors. For this reason, an attempt was made to test the model with four factors.
	Table \ref{tab21} shows the factor loadings for the four-factor model. Table \ref{tab22} shows common variances for primary variables and factors. As in the model with three factors, it is still not possible to indicate which factor carries most of the variance of the primary variables $ x_1 $, $ x_6 $ and $ x_7 $.
	
	After the Varimax rotation, new values of factor loadings were obtained for the four factors (Table \ref{tab23}). When analyzing Table \ref{tab24} containing the level of common variance between the primary variables and the factors, a clear improvement was noticed. After rotation, the first factor carries most of the variances of the primary variables $ x_2 $, $ x_3 $ and $ x_6 $. The second factor represents most of the variances of the primary variables $ x_4 $ and $ x_5 $. The third factor carries most of the variance of the primary variable $ x_7 $, and the fourth factor is representative of the primary variable $ x_1 $.
	
	Row vectors from Table \ref{tab21} (before rotation) and from Table \ref{tab23} (after rotation) can also be compared graphically. Although it is not possible to graphically represent vectors in a four-dimensional space, it is possible to show them by projecting the vectors onto a two-dimensional space.
	Figure \ref{fig5} shows an example of projecting all row vectors from Table \ref{tab21} (before rotation - blue lines) and from Table \ref{tab23} (after rotation - orange lines) onto the $ x_3 \times x_4 $ plane, formed by the $ x_3 $ and $ x_4 $ axes.
	Before the rotation, the third and fourth coordinates in the first and last row vector in Table \ref{tab21} (two pairs of numbers $ [0.595,0.555] $ and $ [0.696, -0.670] $ respectively) had similar values in terms of the modul.
	It is manifested in the fact that in Fig. 5 the line segments which represent the variable $ x_1 $ and $ x_7 $ are distant from the axis of the coordinate system. It can be observed that after the rotation the line segments representing the variables $ x_1 $ and $ x_7 $ became clearly close to the axis of the coordinate system. This means that there are two different factors that represent most of the variances of the variables $ x_1 $ and $ x_7 $.  The above observation is consistent with the conclusions presented above after the analysis of Table \ref{tab24}.
	
	\begin{table}[t]
		\centering
		\caption{The matrix of factor loadings for a four-factor model after Varimax rotation}\label{tab23}
		\fontsize{8}{12}\selectfont{
			\begin{tabular}{ c||c|c|c|c||c| }
				&$ F_1 $&$ F_2 $&$ F_3 $&$ F_4 $&Communality\\ \hline \hline
				$x_1$&-0.129&-0.143&-0.030&0.902&$85.06\%$\\ \hline
				$x_2$&0.956&-0.032&0.051&-0.136&$93.69\%$\\ \hline
				$x_3$&0.853&-0.149&-0.065&-0.324&$85.92\%$\\ \hline
				$x_4$&0.017&0.742&-0.031&-0.177&$58.30\%$\\ \hline
				$x_5$&-0.040&0.840&0.052&0.047&$71.18\%$\\ \hline
				$x_6$&0.733&0.258&0.151&0.319&$72.80\%$\\ \hline
				$x_7$&0.038&0.012&0.992&-0.021&$98.65\%$\\ \hline
			\end{tabular} \\[0.5cm]}
		\caption{The matrix of common variances for a four-factor model after Varimax rotation}\label{tab24}
		\fontsize{8}{12}\selectfont{
			\begin{tabular}{ c||c|c|c|c||c| }	
				&$ F_1 $&$ F_2 $&$ F_3 $&$ F_4 $&Communality\\ \hline \hline
				$x_1$&$1.66\%$&$2.03\%$&$0.09\%$&$81.28\%$&$85.06\%$\\ \hline
				$x_2$&$91.48\%$&$0.10\%$&$0.26\%$&$1.86\%$&$93.69\%$\\ \hline
				$x_3$&$72.77\%$&$2.22\%$&$0.42\%$&$10.51\%$&$85.92\%$\\ \hline
				$x_4$&$0.03\%$&$55.03\%$&$0.09\%$&$3.15\%$&$58.30\%$\\ \hline
				$x_5$&$0.16\%$&$70.53\%$&$0.27\%$&$0.22\%$&$71.18\%$\\ \hline
				$x_6$&$53.70\%$&$6.67\%$&$2.28\%$&$10.16\%$&$72.80\%$\\ \hline
				$x_7$&$0.15\%$&$0.01\%$&$98.44\%$&$0.04\%$&$98.65\%$\\ \hline
		\end{tabular}}
	\end{table}
	
	\begin{table}[t]
		\centering
		\caption{Algorithm for determining the number of factors/components}\label{tab25}
		\fontsize{8}{12}\selectfont{
			\begin{tabular}{r|l|c} \hline
				&\textbf{The steps of the algorith}m&\textbf{Comment}\\ \hline \hline
				\textbf{Input:}&Matrix $ \Lambda_{n\times n} $&Eq. (\ref{r24})\\
				&Matrix $ U_{n\times n} $&Eq. (\ref{r25})\\
				&$ \varepsilon $& Threshold variance\\ \hline
				\textbf{Output:}&NoF & No. of factors/components\\ \hline 
				\textbf{Begin}&&\\
				(01)&$ NoF:=0 $&\\
				(02)&$ MinVar:=0 $& Min. variance\\
				(03)&$ S:=\sqrt{\Lambda} $&Eq. (\ref{r26})\\
				(04)&$ L:=U\cdot S $&Eq. (\ref{r27})\\
				(05)&For $ i:=1 $ to $ n $ do $ C_i:=0 $& Common variances\\
				(06)&$ i:=0 $&\\
				(07)&Do&\\
				(08)&\quad \quad $ i:=i+1 $&\\
				(09)&\quad \quad For $ j:=1 $ to $ n $ do $C_j:=C_j+L_{ji}^2 $&\\
				(10)&\quad \quad $nrVar:= 0$&\\
				(11)&\quad \quad $ MinVar:=1 $&\\
				(12)&\quad \quad For $ j:=1 $ to $ n $ do&\\
				(13)&\quad \quad \quad \quad If ($ C_j<MinVar $) then&\\
				(14)&\quad \quad \quad \quad \quad \quad $ nrVar:= j$&\\
				(15)&\quad \quad \quad \quad \quad \quad $ MinVar:=C_j $&\\
				(16)&While ($ MinVar < \varepsilon $)&\\
				(17)&$ NoF:=i $&\\
				\textbf{End}&&\\ \hline
		\end{tabular}}
	\end{table}	
	
	\section{Common algorithm for determining the number of principal components in PCA and factors in FA}\label{selectAll}
	Subsection \ref{selectCommon}  discusses the problem of determining the appropriate number of principal components due to the need to represent most of the variances of individual primary  variables.
	It was shown there that there is an algorithm for determining the appropriate number of principal components. However, it was found that the suggested algorithm would have too much computational complexity, both in terms of time and memory. The unjustified increase in time complexity would result from the necessity to calculate $ n^2 $ correlation coefficients between $ n $ primary variables and $ n $ principal components.
	On the other hand, an unjustified increase in memory complexity would result from the necessity to use all principal components for the calculation of appropriate correlation coefficients, and not only those that are ultimately necessary to represent most of the variances of the primary variables.
	
	Similarly, subsection \ref{selectInFA} deals with the problem of determining the appropriate number of factors in factor analysis, due to the need to represent most of the variances of individual primary variables by an appropriate factor model. The subsection \ref{selectInFA} mentioned here also suggests that there may be an appropriate algorithm for finding the appropriate number of factors. However, in this case, the proposed algorithm would not need significantly more time and memory complexity.
	
	In subsection \ref{vectFA}, it was found that both principal component analysis and FA share a common vector interpretation. As a result, a version of the algorithm for finding the appropriate number of factors representing most of the variances of primary variables in FA can also be used to find the appropriate number of principal components representing most of the variances of primary variables in PCA. And since this version of the algorithm does not generate greater computational complexity, its use in principal component analysis will also not require greater computational complexity.
	
	A common algorithm for determining the appropriate number of principal components in PCA, as well as determining the appropriate number of factors in FA, is presented in Table \ref{tab25}. The algorithm refers to some common elements found in both PCA and FA. In particular, it uses the diagonal matrix of eigenvalues $ \Lambda $ described by the formula (\ref{r24}) and the matrix of eigenvectors $ U $ (\ref{r25}), which is obtained as a result of solving the eigenproblem  for the matrix of correlation coefficients. The algorithm also uses other variables, the interpretation of which is as follows:
	\begin{itemize}
		\item $ No $F -- a positive integer, obtained as a result of the algorithm's operation, counts the principal components or factors significant from the point of view of representing most of the variances of individual primary variables.
		\item $ \varepsilon $ -- a floating point number greater than $ 0.5 $, arbitrarily taken as a reference minimum value of the variance of each of the primary  variables, which should be represented by principal components or factors. For the purposes of this work, the author assumed the value $ \varepsilon = 0.51 $ (i.e. $ 51\% $) in the calculations.
		\item $ C[n] $ -- $ n- $element non-negative floating point vector that contains variances of individual primary variables represented by $ i $ principal components or $ i $ factors.
		\item $ MinVar $ -- value of the minimum element in the $ C $ array obtained in the $ i-th $ iteration of the loop (12) - (15),
		\item $ nrVar $ -- number of the $ C $ array element containing the smallest variance in the $ i-th $ iteration of the loop (12) - (15).
	\end{itemize}
	Due to the equivalence of the factor loadings matrix and the matrix of correlation coefficients between primary variables and principal components, the presented algorithm is universal. It can be used both in principal component analysis as well as in FA. The presented algorithm also has much lower computational complexity than the algorithm suggested in subsection \ref{selectInFA}, as it does not require multiple computation of correlation coefficients between primary variables and principal components.
	
	\begin{table}[t]
		\centering
		\caption{Modification of the PCA algorithm, taking into account the new criterion for determining the number of principal components}\label{tab26}
		\fontsize{8}{12}\selectfont{
			\begin{tabular}{r|l|c} \hline	
				&\textbf{The steps of the algorith}m&\textbf{Comment}\\ \hline \hline
				\textbf{Input:}&$ X_{m\times n} $&Data matrix (\ref{r20})\\
				&$ \varepsilon $&Threshold variance\\ \hline
				\textbf{Output:}&NoF&No. of PC\\
				&$ P_{m\times NoF} $&Matrix of PC\\ \hline
				\textbf{Begin}&&\\
				(01)&For all the columns of matrix $ X $ find their averages&Eq. (\ref{r1}) \\
				(02)&For the columns of matrix $  X$, &\\
				&find the matrix of their random components $  x$& Eq. (\ref{r2}) \\
				(03)&For all columns of $ X $ find their standard deviations&Eq. (\ref{r6}) \\
				(04)&Standardize the columns of matrix $x$&Eq. (\ref{r7})\\
				(05)&For matrix $ x $, find the matrix of correlation coefficients $ R $&Eq. (\ref{r10}) \\
				(06)&Solve the eigenproblem for the matrix $ R $:&\\
				&\quad \quad -- Find the matrix $ \Lambda $ & Eq. (\ref{r24}) \\
				&\quad \quad -- Find the matrix $ U $&Eq. (\ref{r25}) \\
				(07)&Find the matrix $ S $.&Eq. (\ref{r26}) \\
				(08)&Find the matrix of factor loadings $ L $.&Eq. (\ref{r27}) \\
				(09)&For given $ \varepsilon $ and matrix $ L $,&\\
				&find $ NoF $ which is the final number of PC&Table (\ref{tab25}\\
				(10)&$ k := NoF $&\\
				(11)&Reduce matrix $ U $ to the first $ k $ columns: $ U_{n\times n}\rightarrow U_{n\times k} $&\\
				(12)&Find the matrix of principal components $ P_{m\times k} $: &\\
				&$ P_{m\times k} := x_{m\times n} \cdot U_{n \times k} $&Eq. (\ref{r15}) or Eq. (\ref{r49}) \\ 
				\textbf{End}&&\\ \hline
		\end{tabular}}
	\end{table}	
	
	\subsection{Modification of the principal components analysis algorithm}
	The criteria for determining the number of factors (or principal components) described in subsection \ref{selectCommon} are blind to the variance values of single primary variables. It may happen that the factors (or principal components) determined on the basis of these criteria do not represent most of the variance of some of the primary variables. On the other hand, the criterion presented in section \ref{selectAll} avoids this deficit. The algorithm using the above criterion allows for a more reliable way of determining the number of factors (principal components). An example of the use of this algorithm in FA is presented in section \ref{selectInFA}. Here, the discussed algorithm will be used to modify the PCA in order to enable the determination of the optimal number of principal components. The modified version of the PCA algorithm is presented in Table \ref{tab26}.
	
	\subsection{Examples of determining the number of factors/components}\label{examples}
	In determining the appropriate number of principal components in PCA or factors in FA, different criteria may lead to the recommendation of a different number of principal components or factors, and thus may lead to inconsistent results. On the one hand, obtaining identical results is not excluded. On the other hand, recommendations for the number of factors obtained by different methods may be different. First of all, it may happen that some criteria may lead to a recommendation that is unsatisfactory from the point of view of the representation of most of the variances of the primary variables. Four examples will be shown in this subsection which will confirm the necessity to apply the criterion presented in section \ref{selectAll}.
	
	\begin{table}[t]
		\centering
		\caption{The percentage of variances explained by the successive factors for Dataset No. 2}\label{tab27}
		\fontsize{8}{12}\selectfont{
			\begin{tabular}{c|c|c|c|c} \hline 
				\multirow{2}{*}{No.} & \multirow{2}{*}{Eigenvalue} & Cumulative & Percentage of variance & Cumulative \\ 
				& & eigenvalues & explained by each PC & percentage of variance \\ \hline \hline
				1&3.912&3.912&$43.5\%$&$43.5\%$\\ \hline
				2&1.923&5.835&$21.4\%$&$64.8\%$\\ \hline
				3&1.697&7.532&$18.9\%$&$83.7\%$\\ \hline
				4&0.910&8.442&$10.1\%$&$93.8\%$\\ \hline
				5&0.293&8.736&$3.3\%$&$97.1\%$\\ \hline
				6&0.143&8.878&$1.6\%$&$98.6\%$\\ \hline
				7&0.063&8.941&$0.7\%$&$99.3\%$\\ \hline
				8&0.045&8.985&$0.5\%$&$99.8\%$\\ \hline
				9&0.015&9.000&$0.2\%$&$100\%$\\ \hline
			\end{tabular} \\[0.5cm]}
		\caption{Minimum variance (MinVar) and mean variance (AverVar) reproduced by successive factors (Dataset No. 2)}\label{tab28}
		\fontsize{8}{12}\selectfont{
			\begin{tabular}{ c||c|c|c|c|c|c|c|c|c| }
				No. of factors&1&2&3&4&5&6&7&8&9\\ \hline \hline
				EigVal&$43.5\%$&$21.4\%$&$18.9\%$&$10.1\%$&$3.3\%$&$1.6\%$&$0.7\%$&$0.5\%$&$0.2\%$\\ \hline
				MinVar&$0.8\%$&$7.3\%$&$18.8\%$&$87.4\%$&$90.4\%$&$96.9\%$&$98.3\%$&$99.3\%$&$100\%$\\ \hline
				AverVar&$43.5\%$&$64.8\%$&$83.7\%$&$93.8\%$&$97.1\%$&$98.7\%$&$99.3\%$&$99.8\%$&$100\%$\\ \hline
				NrMinVar&1&2&3&1&6&4&8&5&2
		\end{tabular}}
	\end{table}
	\begin{figure}[h!]
		\centering
		\includegraphics[width=0.65\textwidth]{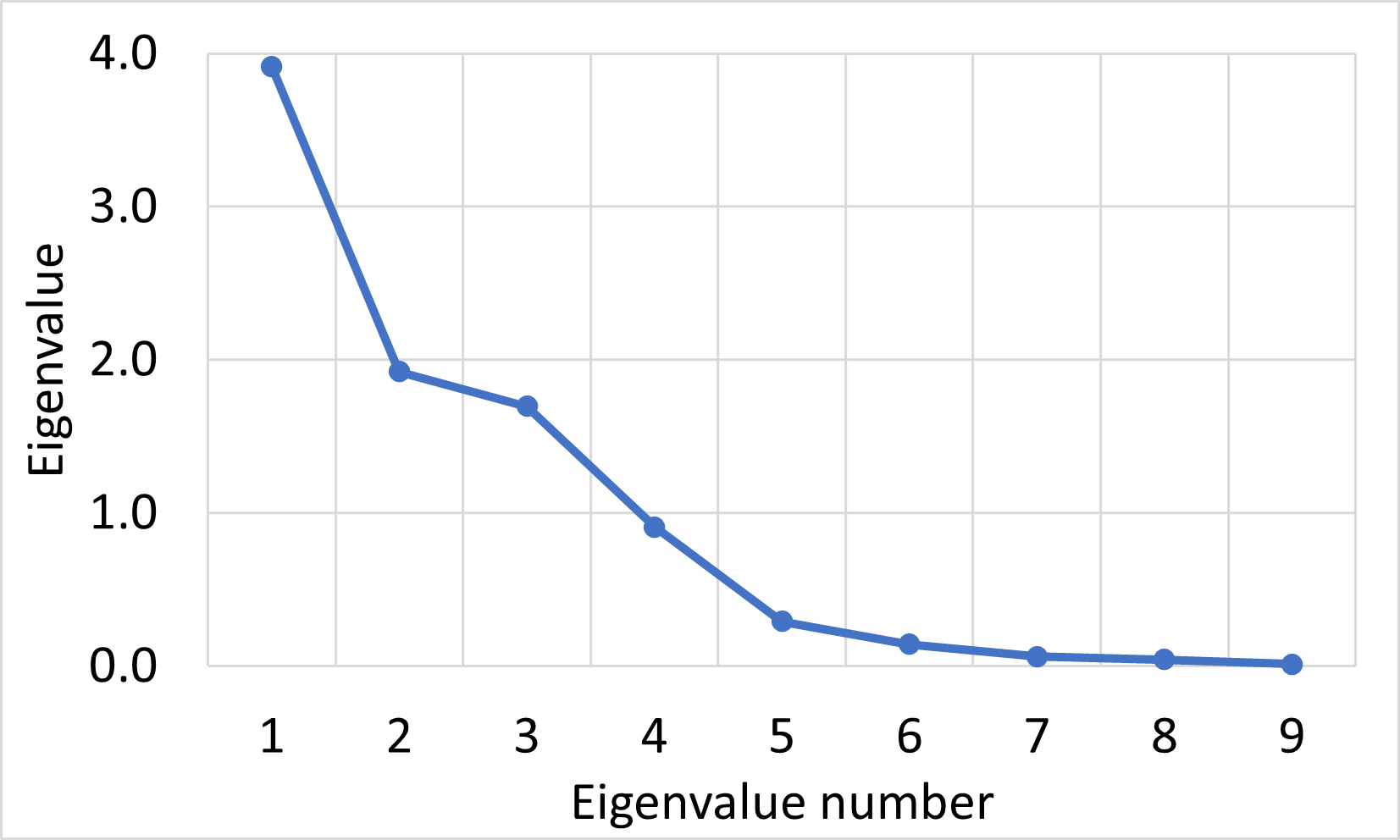}
		\caption{Scree plot for Dataset No. 2}\label{fig6}
	\end{figure}
	\begin{figure}[h!]
		\centering
		\includegraphics[width=0.65\textwidth]{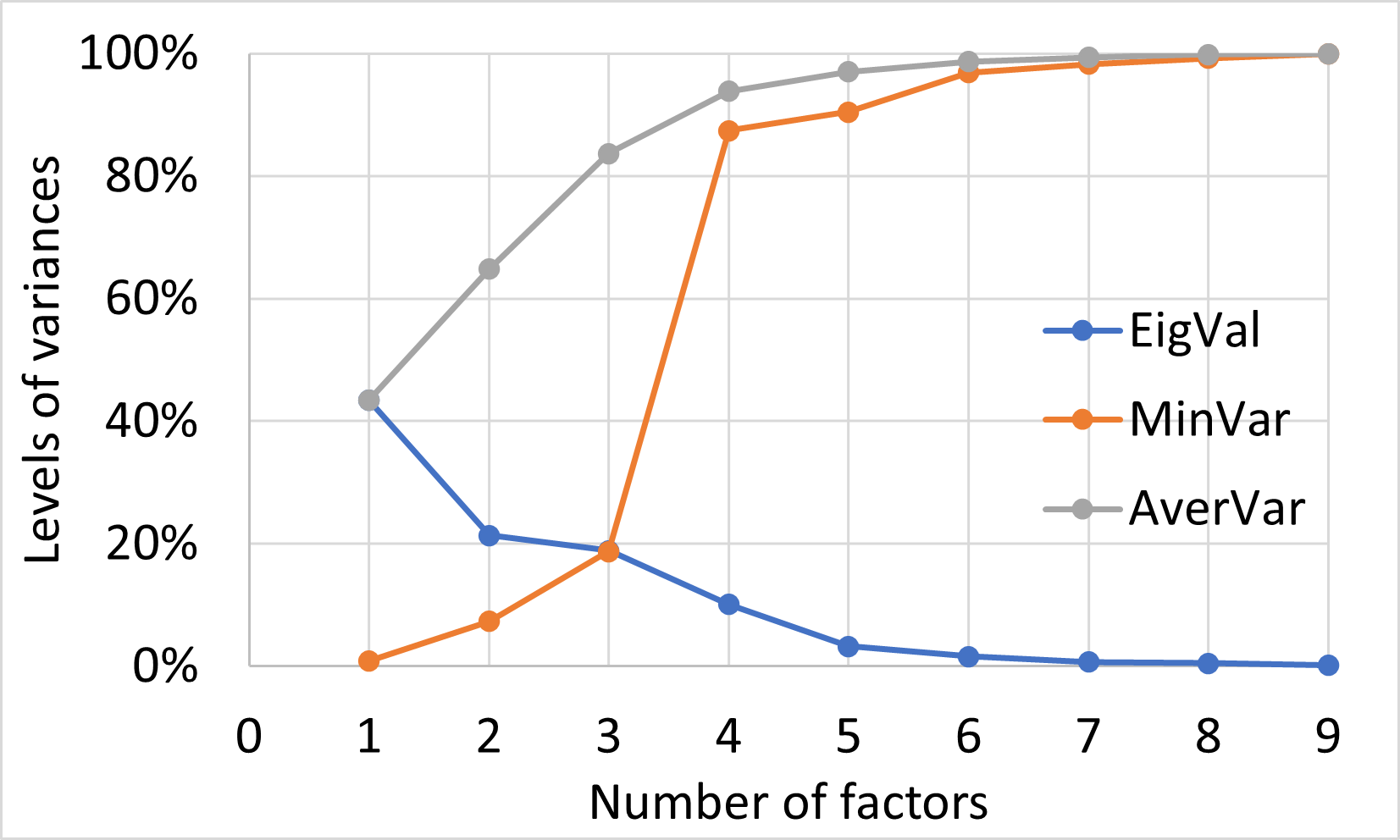}
		\caption{Minimum variance (MinVar) and mean variance (AverVar) reproduced by successive factors  (Dataset No. 2)}\label{fig7}
	\end{figure}
	
	\subsubsection{Dataset No. 2}
	A dataset known as ''Houses Data'' was used to test the effectiveness of the algorithms for determining the number of factors or principal components. This dataset is used in the book \cite{Larose2006} and also in the article \cite{Pace1997}. A link to the location of the dataset \cite{Pace1999} is given in \cite{Larose2006}. The dataset contains nine variables that have been measured $ 41280 $ times. The first variable was adopted as the dependent variable. This variable was modeled using the remaining eight variables. Principal component analysis for these eight variables was performed in \cite{Larose2006}. However, since all variables are correlated, in this article the analysis was performed for all nine variables. The matrix of correlation coefficients and the matrix of determination coefficients for the discussed data set are presented in \cite{Gniazdowski2018}. Table \ref{tab27} shows the variance distribution explained by the following factors. When analyzing the table, it can be noticed that the choice of three factors will explain slightly over $ 83\% $ of the variance of the primary variables. The Kaiser criterion also suggests the choice of three factors. On the other hand, in the scree plot (Fig. \ref{fig6}), the four eigenvalues are above the ''elbow'' on the slope of the scree. The analysis of Table \ref{tab27} and Figure \ref{fig7} shows that the choice of three factors will explain only $ 18.75\% $ of the variance of the variable $ x_3 $. Therefore, four factors must be selected that explain more than $ 87\% $ of the variance of each primary variable.
	
	\begin{table}[t]
		\centering
		\caption{The percentage of variances explained by the successive factors for Dataset No. 3}\label{tab29}
		\fontsize{8}{12}\selectfont{
			\begin{tabular}{c|c|c|c|c} \hline 
				\multirow{2}{*}{No.} & \multirow{2}{*}{Eigenvalue} & Cumulative & Percentage of variance & Cumulative \\ 
				& & eigenvalues & explained by each PC & percentage of variance \\ \hline \hline
				1&58.3&58.3&$47.4\%$&$47.4\%$ \\ \hline
				2&19.5&77.8&$15.9\%$&$63.2\%$ \\ \hline
				3&17.9&95.7&$14.6\%$&$77.8\%$ \\ \hline
				4&8.5&104.2&$6.9\%$&$84.7\%$ \\ \hline
				5&7.3&111.5&$6.0\%$&$90.7\%$ \\ \hline
				6&4.2&115.7&$3.4\%$&$94.1\%$ \\ \hline
				7&2.2&117.9&$1.8\%$&$95.9\%$ \\ \hline
				8&0.8&118.7&$0.6\%$&$96.5\%$ \\ \hline
				9&0.5&119.2&$0.4\%$&$96.9\%$ \\ \hline
				$\cdots$&$\cdots$&$\cdots$&$\cdots$&$\cdots$ \\ \hline
			\end{tabular} \\[0.5cm]}
		\caption{Minimum variance (MinVar) and mean variance (AverVar) reproduced by successive factors (Dataset No. 3)}\label{tab30}
		\fontsize{8}{12}\selectfont{
			\begin{tabular}{ c||c|c|c|c|c|c|c|c|c| }
				No. of factors&1&2&3&4&5&6&7&8&$\cdots$\\ \hline 
				EigVal&$47.4\%$&$15.9\%$&$14.6\%$&$6.9\%$&$6.0\%$&$3.4\%$&$1.8\%$&$0.6\%$&$\cdots$ \\ \hline
				MinVar&$3.9\%$&$4.0\%$&$12.5\%$&$13.3\%$&$14.3\%$&$33.2\%$&$52.0\%$&$60.1\%$&$\cdots$ \\ \hline
				AverVar&$47.4\%$&$63.2\%$&$77.8\%$&$84.7\%$&$90.7\%$&$94.1\%$&$95.9\%$&$96.5\%$&$\cdots$ \\ \hline
				NrMinVar&81&81&8&8&8&11&4&11&$\cdots$ \\ \hline
		\end{tabular}}
	\end{table}
	\begin{figure}[h!]
		\centering
		\includegraphics[width=0.65\textwidth]{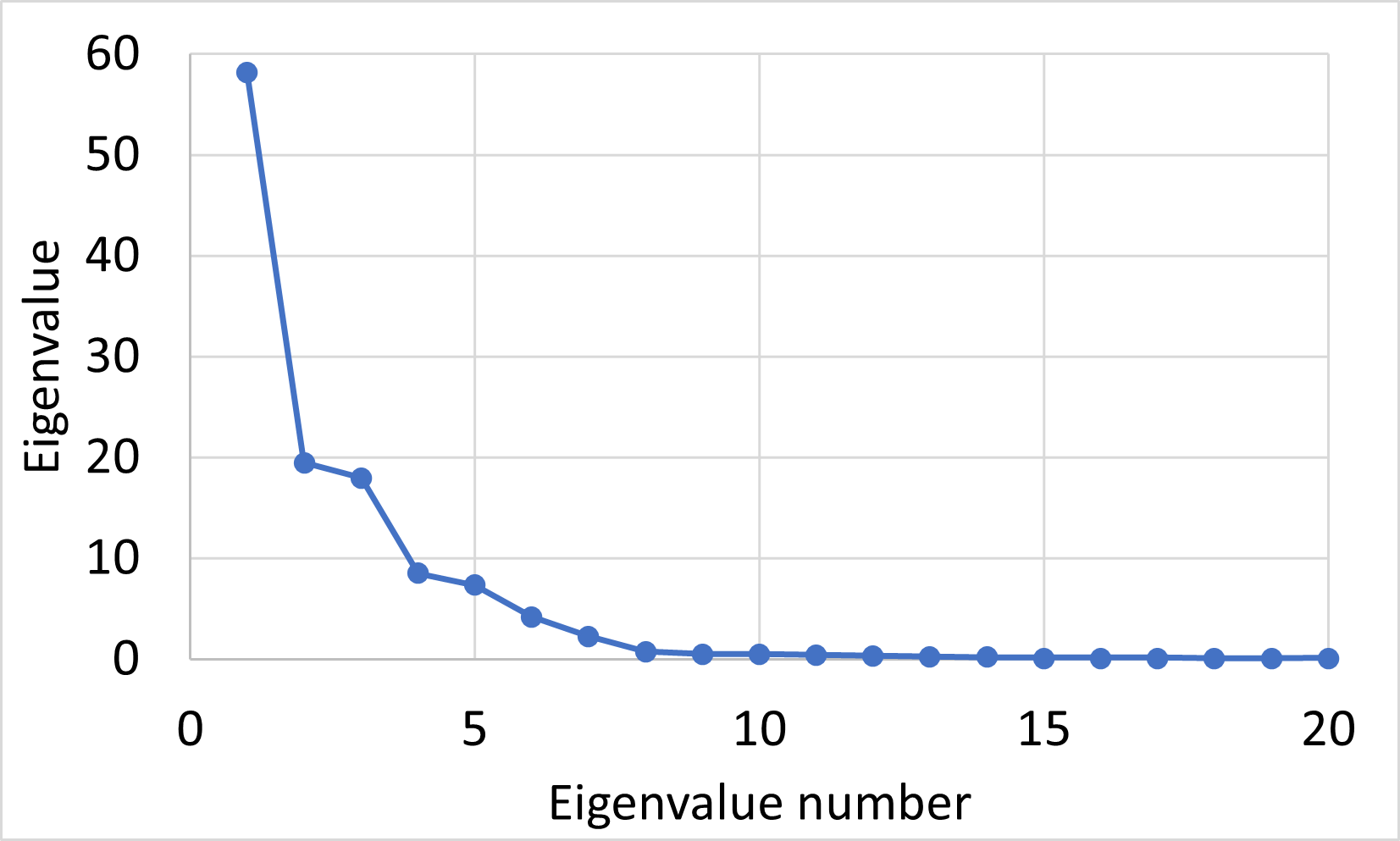}
		\caption{Scree plot for Dataset No. 3}\label{fig8}
	\end{figure}
	\begin{figure}[h!]
		\centering
		\includegraphics[width=0.65\textwidth]{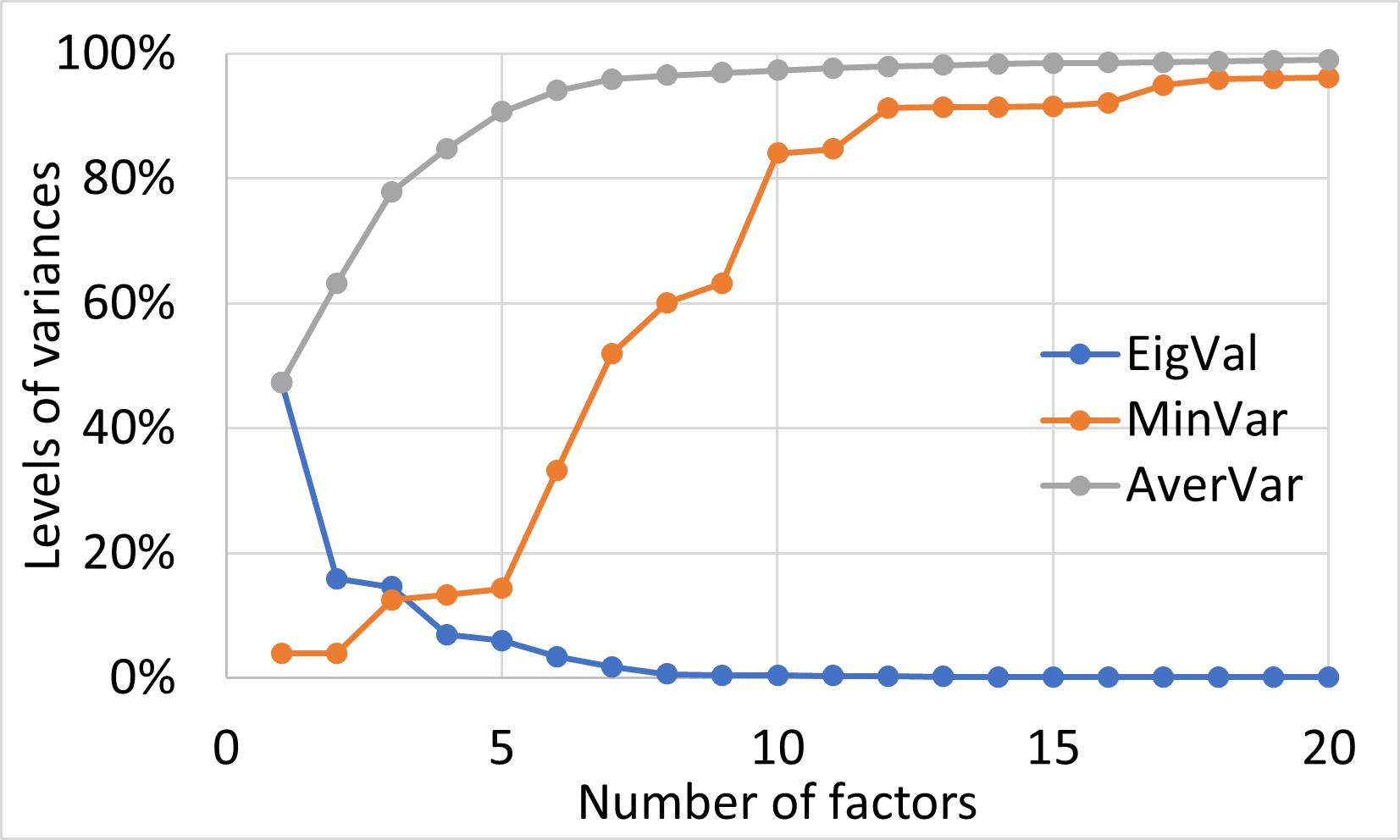}
		\caption{Minimum variance (MinVar) and mean variance (AverVar) reproduced by successive factors  (Dataset No. 3)}\label{fig9}
	\end{figure}
	
	\subsubsection{Dataset No. 3}
	As another example, the dataset used in \cite{Poelstra2015} and shared in \cite{Poelstra2015a} will be shown. The dataset contains $ 123 $ correlated random variables, each of which has been measured $ 14504 $ times. Table \ref{tab29} contains the first nine lines describing the distribution of variances explained by the successive factors. The last column of the table shows that the choice of four factors will explain more than $ 84\% $ of the variance of the primary variables. The Kaiser criterion suggests the use of seven factors. On the other hand, in Figure \ref{fig8}, there are three so-called ''Elbows''. This fact does not make analysis easier. The final results are not unequivocal. On the other hand, analysis of Table \ref{tab30} and Figure \ref{fig9} shows that selecting the seven factors will represent most of the variance of each of the primary variables (see MinVar). This clearly suggests that seven factors (components) should be selected for both factor analysis and principal component analysis.
	
	\begin{table}[t]
		\centering
		\caption{The percentage of variances explained by the successive factors for Dataset No. 4}\label{tab31}
		\fontsize{8}{12}\selectfont{
			\begin{tabular}{c|c|c|c|c} \hline 
				\multirow{2}{*}{No.} & \multirow{2}{*}{Eigenvalue} & Cumulative & Percentage of variance & Cumulative \\ 
				& & eigenvalues & explained by each PC & percentage of variance \\ \hline \hline
				1&1.812&1.812&$25.9\%$&$25.9\%$\\ \hline
				2&1.697&3.508&$24.2\%$&$50.1\%$\\ \hline
				3&1.481&4.990&$21.2\%$&$71.3\%$\\ \hline
				4&1.000&5.990&$14.3\%$&$85.6\%$\\ \hline
				5&0.813&6.803&$11.6\%$&$97.2\%$\\ \hline
				6&0.189&6.992&$2.7\%$&$99.9\%$\\ \hline
				7&0.008&7.000&$0.1\%$&$100\%$\\ \hline
			\end{tabular} \\[0.5cm]}
		\caption{Minimum variance (MinVar) and mean variance (AverVar) reproduced by successive factors (Dataset No. 4)}\label{tab32}
		\fontsize{8}{12}\selectfont{
			\begin{tabular}{ c||c|c|c|c|c|c|c| }
				No. of factors&1&2&3&4&5&6&7\\ \hline 
				EigVal&$25.88\%$&$24.24\%$&$21.16\%$&$14.29\%$&$11.62\%$&$2.70\%$&$0.11\%$\\ \hline
				MinVar&$0.00\%$&$0.06\%$&$0.06\%$&$44.43\%$&$90.54\%$&$99.73\%$&$100\%$\\ \hline
				AverVar&$25.88\%$&$50.12\%$&$71.28\%$&$85.57\%$&$97.19\%$&$99.89\%$&$100\%$\\ \hline
				NrMinVar&3&1&1&2&7&3&1\\ \hline
		\end{tabular}}
	\end{table}
	\begin{figure}[h!]
		\centering
		\includegraphics[width=0.65\textwidth]{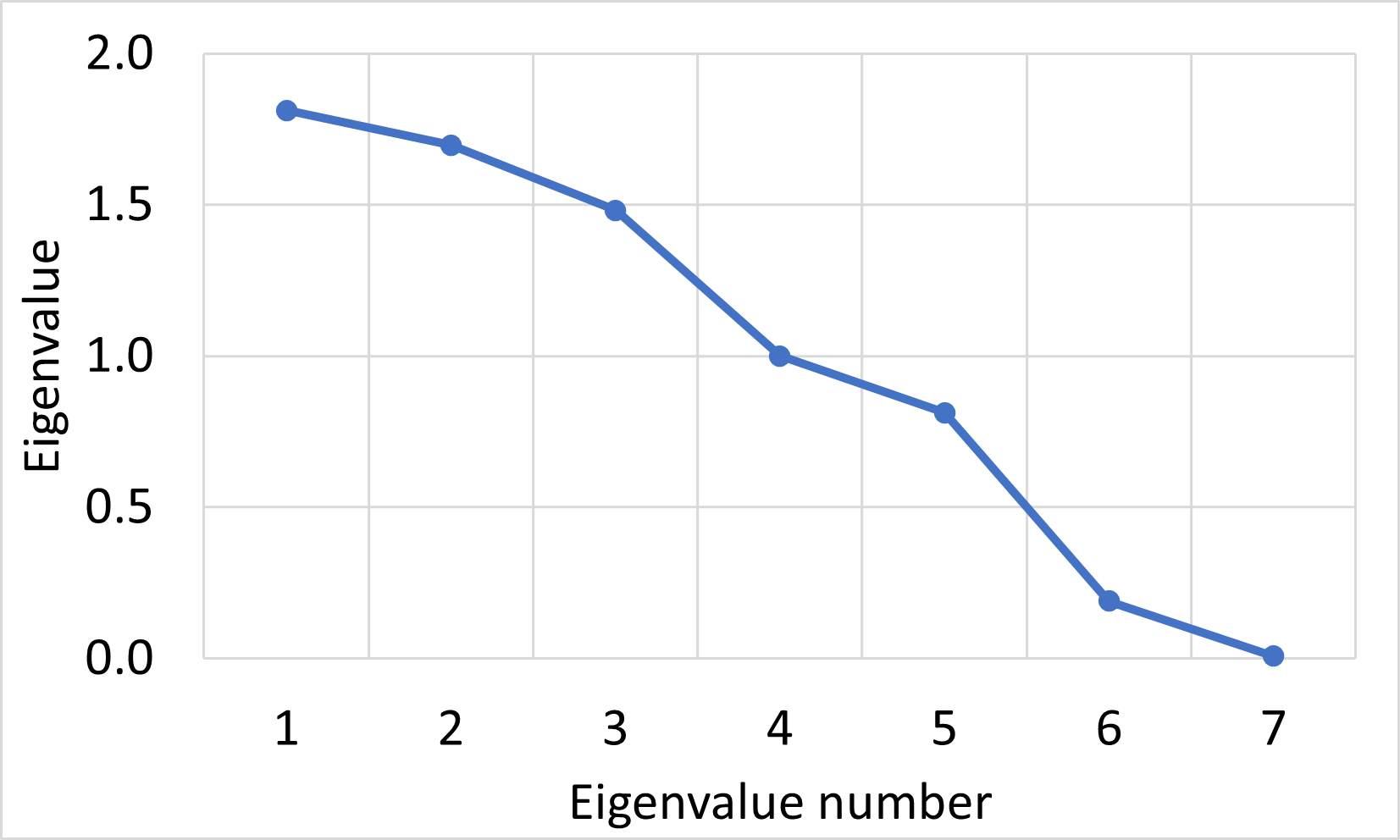}
		\caption{Scree plot for Dataset No. 4}\label{fig10}
	\end{figure}
	\begin{figure}[h!]
		\centering
		\includegraphics[width=0.65\textwidth]{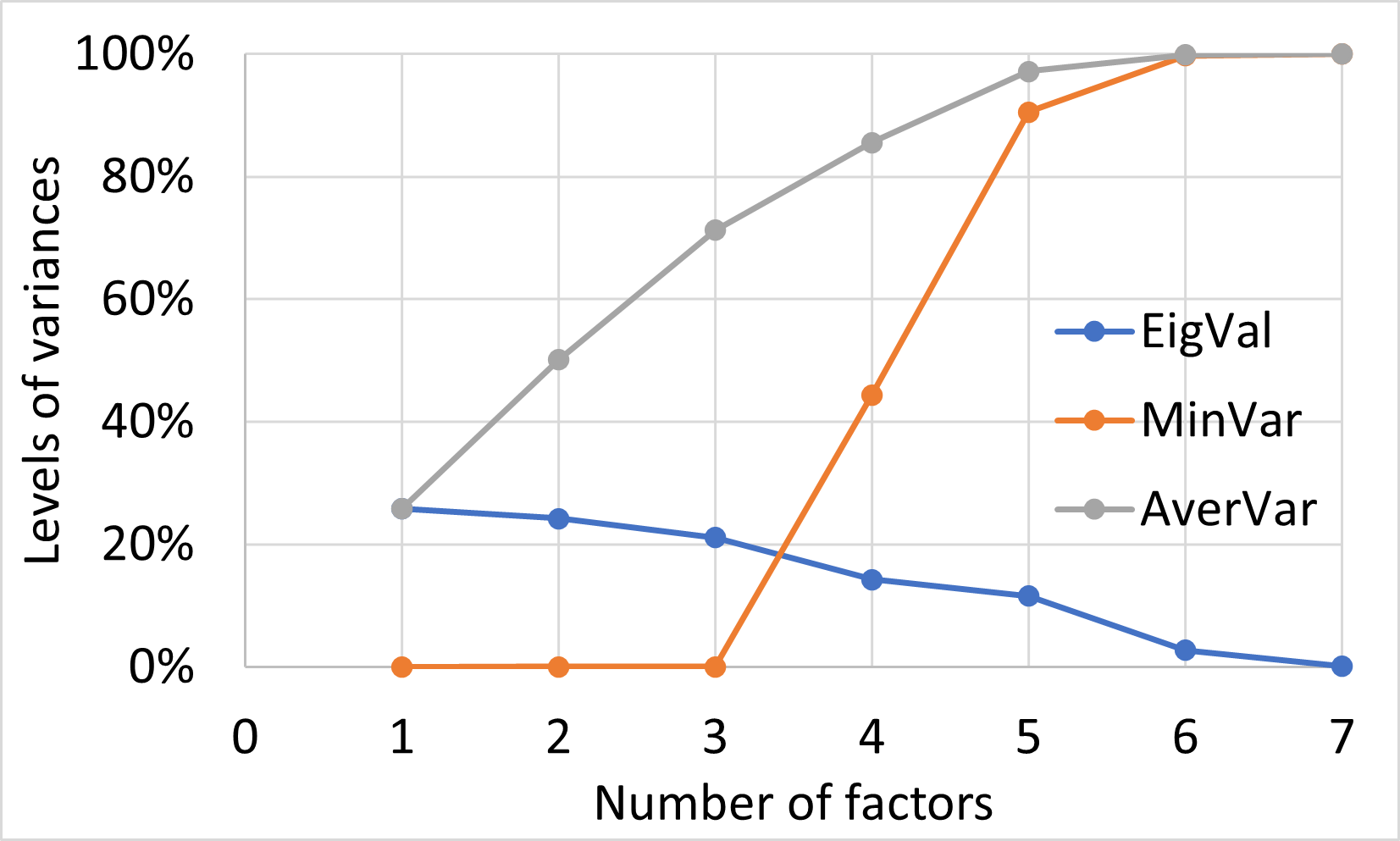}
		\caption{Minimum variance (MinVar) and mean variance (AverVar) reproduced by successive factors  (Dataset No. 4)}\label{fig11}
	\end{figure}
	
	\subsubsection{Dataset No. 4}\label{examp4}
	Another data set was prepared at the Ship Hydromechanics Laboratory at the Maritime and Transport Technology Department of the Delft University of Technology. Its content was made available by the UCI Machine Learning Repository \cite{Gerritsma2013}. The analyzed dataset contains seven random variables measured $ 308 $ times.
	
\begin{figure}[h!]
		\centering
		\includegraphics[width=0.65\textwidth]{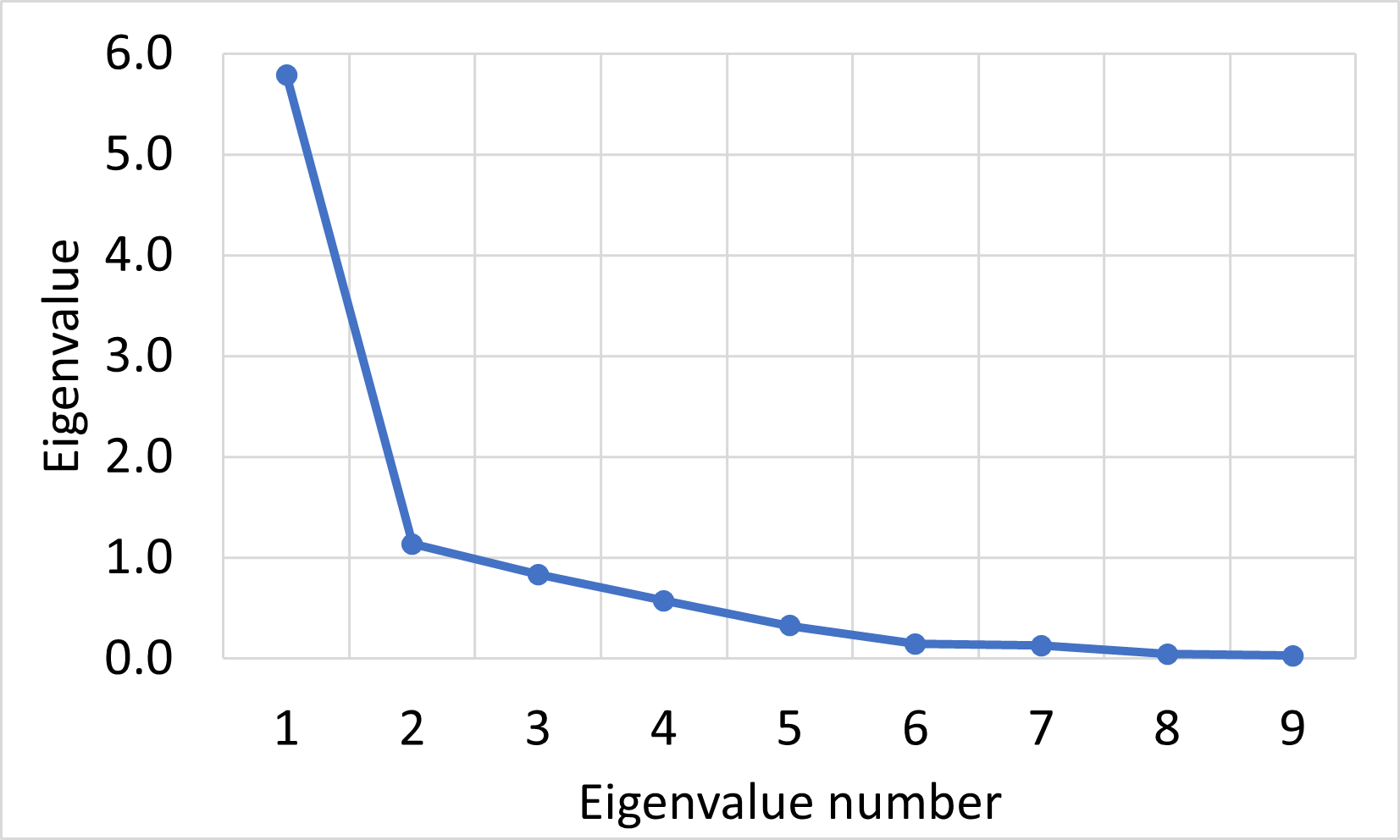}
		\caption{Scree plot for Dataset No. 5}\label{fig12}
	\end{figure}

	In the case of determining the appropriate number of factors for the analyzed data set, it should be stated that the criterion of the scree plot is not adequate here, because the graph does not indicate two phases before ''elbow'' and after ''elbow''. There is no ''elbow'' in Figure \ref{fig12}. On the other hand, the criterion of half the number of primary variables suggests the selection of three factors, and the Kaiser criterion -- four. The last classic criterion, i.e. the criterion of explained variance (Table \ref{tab33}), suggests three factors ($ 71.3\% $ of variance) or four factors ($ 85.6\% $ of variance) depending on the accepted minimum threshold of explained variance. As shown above, this criterion only informs about the average level of reproduction of the variance of all primary variables.Using this criterion, it is possible that the variance of individual primary variables may not be sufficiently reproduced.
	
\begin{figure}[h!]
		\centering
		\includegraphics[width=0.65\textwidth]{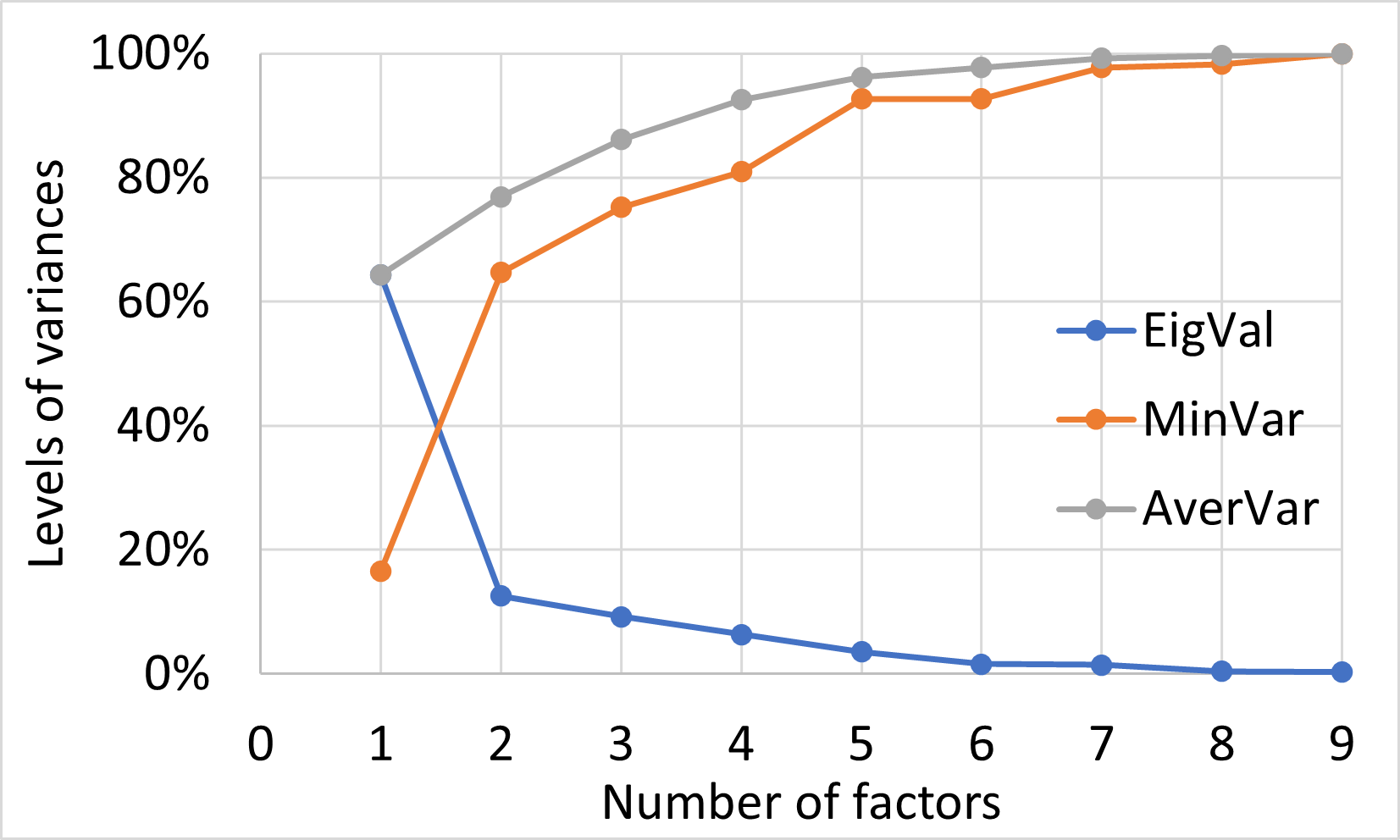}
		\caption{Minimum variance (MinVar) and mean variance (AverVar) reproduced by successive factors  (Dataset No. 5)}\label{fig13}
	\end{figure}

	And it really is. The analysis of Table \ref{tab34} and Figure \ref{fig13} shows that for three factors, at least the variance of the first primary variable x will be insufficiently reproduced. For three factors the level of reproduction of variance $ x_1 $ will be less than $ 1\% $. From the point of view of the possibility of reproducing most of the variances of single primary variables, also four factors are not enough. With four factors, the variance $ x_2 $ will be reproduced in $ 44.4\% $. Only five factors will reproduce most of the variance of all single primary variables.
	
	\begin{table}[t]
		\centering
		\caption{The percentage of variances explained by the successive factors for Dataset No. 5}\label{tab33}
		\fontsize{8}{12}\selectfont{
			\begin{tabular}{c|c|c|c|c} \hline 
				\multirow{2}{*}{No.} & \multirow{2}{*}{Eigenvalue} & Cumulative & Percentage of variance & Cumulative \\ 
				& & eigenvalues & explained by each PC & percentage of variance \\ \hline \hline
				1&5.787&5.787&$64.3\%$&$64.3\%$\\ \hline
				2&1.135&6.922&$12.6\%$&$76.9\%$\\ \hline
				3&0.833&7.756&$9.3\%$&$86.2\%$\\ \hline
				4&0.575&8.330&$6.4\%$&$92.6\%$\\ \hline
				5&0.325&8.655&$3.6\%$&$96.2\%$\\ \hline
				6&0.145&8.800&$1.6\%$&$97.8\%$\\ \hline
				7&0.129&8.929&$1.4\%$&$99.2\%$\\ \hline
				8&0.043&8.972&$0.5\%$&$99.7\%$\\ \hline
				9&0.029&9.000&$0.3\%$&$100\%$\\ \hline
			\end{tabular} \\[0.5cm]}
		\caption{Minimum variance (MinVar) and mean variance (AverVar) reproduced by successive factors (Dataset No. 5)}\label{tab34}
		\fontsize{8}{12}\selectfont{
			\begin{tabular}{ c||c|c|c|c|c|c|c|c|c| }
				No. of factors&1&2&3&4&5&6&7&8&9\\ \hline
				EigVal&$64.3\%$&$12.6\%$&$9.3\%$&$6.4\%$&$3.6\%$&$1.6\%$&$1.4\%$&$0.5\%$&$0.3\% $ \\ \hline 
				MinVar&$16.6\%$&$64.8\%$&$75.2\%$&$81.0\%$&$92.7\%$&$92.7\%$&$97.8\%$&$98.3\%$&$100\% $ \\ \hline
				AverVar&$64.3\%$&$76.9\%$&$86.2\%$&$92.6\%$&$96.2\%$&$97.8\%$&$99.2\%$&$99.7\%$&$100\% $ \\ \hline
				NrMinVar&6&7&7&3&4&4&2&8&7 \\ \hline
		\end{tabular}}
	\end{table}
	
	\subsubsection{Dataset No. 5}
	The dataset contains the value of the Istanbul Stock Exchange Index along with seven other stock indices. The data was collected between June 5, 2009 and February 22, 2011. The source of the data can be found on the UCI Machine Learning Repository website \cite{Akbilgic2013}. The data table has $ 536 $ rows and $ 10 $ columns. After the date is rejected, $ 9 $ columns remain for analysis.
	
	As in the previous examples, the determination of the appropriate number of factors was made in the context of the four classical criteria and the criterion developed in this article:
	\begin{itemize}
		\item Figure \ref{fig12} shows a scree plot. As there is one point in front of the so-called ''elbow'' on the slope of the scree, the criterion of the scree plot suggests the selection of one factor.
		\item 	The Kaiser criterion suggests two factors because two eigenvalues are greater than one. 
		\item 	Four factors suggest the criterion of half the number of primary variables.
		\item 	Assuming that the average level of the explained variance should be at least $ 80\% $, on the basis of Table \ref{tab33} it can be said that the criterion of the explained variance suggests the selection of three factors.
	\end{itemize}
	Each of the criteria suggests a different solution to the problem of determining the number of factors.
	
	There is still the last criterion, discussed in this article. From Table \ref{tab34} and Figure \ref{fig13}, it can be seen that one factor would explain less than $ 17\% $ of the variance of the primary variable $ x_6 $. Two factors explain the variance of the primary variable $ x_7 $ at almost $ 65\% $. Therefore, from the point of view of the necessity to reproduce most of the variance of each of the primary variables, two factors are sufficient.
	
	\section{Discussion}
	The article attempts to compare the exploratory factor analysis based on principal components with the principal components analysis using the correlation coefficient matrix. Both types of analyzes have a common mathematical core. Among the elements common to both types of analyzes, there are also non-obvious elements. Here an attempt will be made to discuss them. Particular attention will be paid to a common algorithm for determining the number of factors in FA and principal components in PCA.
	
	\begin{table}
		\centering
		\caption{Comparison for significant differences between PCA and FA}\label{tab35}
		\fontsize{7}{10}\selectfont{
			\begin{tabular}{|c|p{5cm}|p{4cm}|} \hline
				&\textbf{Principal component analysis}&\textbf{Factor analysis} \\ \hline \hline
				Essence&The observed correlated random variables are replaced by a set of independent random variables. In essence, this operation is based on the orthogonal rotation of the Cartesian coordinate system. The measured points are projected onto the axes of the new (different from the original) coordinate system and then read in this new coordinate system.&Based on the observations, linear models of the observed correlated primary variables are built with respect to the set of independent standard factors. \\ \hline
				Aim&PCA only aims to recreate the sample used&EFA aims to model the target population\\ \hline
				Algorithm& \begin{itemize}[nosep,leftmargin=10pt] \item Successive eigenvectors become successive columns of the matrix. \item The transposition of the eigenvector matrix is an orthogonal rotation matrix. \item The product of the matrix containing the standardized primary random variables by the transposed rotation matrix gives the matrix of principal components.\end{itemize}& \begin{itemize}[nosep,leftmargin=10pt] \item Successive eigenvectors become successive columns of the matrix. \item The product of the matrix of eigenvectors by the diagonal matrix of the roots of the eigenvalues gives the factor model. \end{itemize}\\ \hline
				Result&Principal components are representatives of primary variables:\newline Single principal component = linear combination of the observed variables&Factor analysis provides models for primary variables:\newline Single observed variable = linear combination of factors (components) + error\\ \hline
				Interpretation&Principal components have no interpretation&Factors are subject to interpretation. \\ \hline
				Ambiguity&The obtained solution is unambiguous.&There are many different solutions. \\ \hline
				Benefits&\begin{itemize}[nosep,leftmargin=10pt] \item From the set of principal members, you can remove those components that have the smallest variance. In this case, most of the information contained in the set of primary variables will be represented by a smaller set of principal components. In addition, such a reduction can also be viewed as a lossy compression of the input data. \item A reduced set of principal components enables more effective data clustering. Clustering in a reduced space is less computationally intensive. \item In the case of reducing the principal components to two, there is a possibility of effective data visualization, as well as the assessment of the possibility of their clustering. \item In the least squares method, the use of principal components instead of primary variables prevents errors resulting from the ill-conditioning of the matrix of a system of normal equations \end{itemize}&Factor analysis models $ n $ random variables against fewer hidden factors. This gives the following possibilities: \begin{itemize}[nosep,leftmargin=10pt] \item Hidden factors can be interpreted (identified). Their correct interpretation makes it possible to explain the random phenomenon represented by the primary variables, and thus to explain the common causes that influence the observed phenomenon. \item Primary variables can be clustered due to their similarity to factors. \item If the factor model is known, and the primary variables are not available, the factor model will enable the Monte Carlo simulation of the primary variables, and then the estimation of their statistical characteristics.
				\end{itemize}\\ \hline
		\end{tabular}}
	\end{table}
	
	\subsection{Differences between the PCA and the FA}
	The differences between principal component analysis and factor analysis are shown in Table \ref{tab35}.
	In the columns of the table, it is possible to analyze the detailed differences between both types of analyzes. It can be noticed that these differences refer to different aspects of both types of analyzes, such as their essence, goals, similarity of algorithms, obtained results, ambiguity of solutions, interpretation of factors, as well as benefits.
	\subsection{Similarities between the PCA and the FA}
	There are also significant similarities between the PCA and the FA. These similarities result from the common mathematical core of both types of analysis. This common core is eigenproblem solving for the matrix of correlation coefficients. Detailed similarities resulting from this core refer to several important aspects, such as the previously discussed geometric interpretation, equivalence of factors before rotation with principal components, methods of determining the number of principal components and factors, or the observed artifact, as well as clustering of primary variables due to their similarity to both principal components and factors.
	
	In subsections \ref{notesOn} and \ref{commonElem} it was stated that both types of analysis (PCA and FA) use common or analogous algorithms. It was found that the common algorithm is eigenproblem solving for the correlation coefficient matrix, and the analogous algorithm is the determination of the number of principal components and factors. After further analysis carried out in this article, and in particular after finding that the matrix of factor loadings before rotation is the same as the matrix of correlation coefficients between primary variables and principal components, it can be concluded that the algorithms included in the PCA and FA, which in subsection \ref{notesOn} were considered analogous, now they can obtain the status of common algorithms, both for the PCA and for the FA. In particular, the algorithm that obtained the status of a common algorithm for both types of analyzes is the algorithm for determining the number of principal components and the number of factors. All of these algorithms, or more broadly the elements that are common to both PCA and FA, will be discussed in detail later in this section. 
	
	\subsubsection{Geometric interpretation of factor analysis}
	The exploratory factor analysis based on principal components was compared with the principal components analysis using the matrix of correlation coefficients. The article \cite{Gniazdowski2017} proposes a geometric interpretation of the principal components analysis in the vector space. It was found that standardized primary variables can be presented as vectors whose components are equal to the correlation coefficients between primary variables and principal components.
	
	An analogous interpretation was proposed for the factor analysis. In this case, the primary variables are also presented as vectors whose components are equal to the factor loadings. It was found that the values of the factor loadings connecting the primary variables with the independent factors are equal to the above-mentioned correlation coefficients between the primary variables and the principal components obtained in the PCA.
	Consequently, the matrix of correlation coefficients between primary variables  and principal components obtained in the principal components analysis is identical to the matrix of factor loadings obtained in the factor analysis.
	This means that the two vector interpretations of the primary variables are not only analogous, but identical.
	
	The consequence of this vector representation is the possibility of using the Pythagorean theorem to describe the behavior of primary variables. On the other hand, the cosines between the individual vectors representing the primary variables are the same as the correlation coefficients between the corresponding fundamental variables.
	
	\subsubsection{Factors before rotation versus standardized principal components}
	The article examines the factor analysis using principal components. In this analysis, the factors before their reduction can be identified with standardized principal components obtained in principal components analysis. More precisely, the factors become variables identical to the first few standardized principal components with the largest variances. This is because:
	\begin{enumerate}
		\item The factor loadings that relate the modeled primary variables to the factors are directly computed with the eigenvalues and eigenvectors estimated for the correlation coefficient matrix. The eigenvalues are equal to the variances of the successive principal components \cite{Gniazdowski2017}. Eigenvectors are used in the transformation (\ref{r27}). This transformation leads to the estimation of the factor loadings.
		\item The factor loadings are the same as the correlation coefficients between the primary variables and the principal components in the PCA (see subsection \ref{geomPCA}). Both the factor squares and the corresponding squares of the correlation coefficients measure the level of common variance between the primary variables and the factors in FA and principal components in PCA.
	\end{enumerate}
	Hence, it is justified to state that the factor analysis before any rotation of factors, models the standardized primary variables using standardized principal components.
	
	\subsubsection{The problem of determining the number of factors/components}
	The criteria for determining the number of principal components and factors described in subsection \ref{selectCommon} were subjected to a detailed critical analysis. 
	Examples of their weaknesses are provided in subsection \ref{examples}:
	\begin{itemize}
		\item The scree plot criterion -- the examples presented in subsection \ref{examples} show that in some situations the scree plot is ambiguous:
		\begin{itemize}
			\item The scree diagram does not show two phases separated by the so-called ''elbow'' (Figures \ref{fig2} and \ref{fig10}),
			\item More than one ''elbow'' can be seen in the scree plot (Figures \ref{fig6} and \ref{fig8}).
		\end{itemize}
		\item Percentage criterion of explained variance - the weakness of this criterion is that it relates to the average variance of the primary variables represented by the selected factors. 	 Depending on the distribution of the obtained eigenvalues, the explained mean variance of the primary variables may be relatively large, and the reconstructed variance of individual primary variables may be negligible (Table \ref{tab32}, Fig. \ref{fig11}).
		\item  Eigenvalue criterion called the Kaiser criterion - the fact that a given factor with an eigenvalue greater than one should have a variance greater than the variance of a single primary variable does not mean that a factor with a variance of less than one will never represent most of the variance of some primary variable. On the other hand, if a factor with an eigenvalue less than one would represent most of the variance of some primary variable, then that factor should not be rejected. It should also be added that this criterion does not consider rotation, which can radically change the situation by assigning significant factor loadings to the non-rejected factor.
		\item  The criterion of half the number of primary variables - in practice, there may be situations in which the mutual correlations between the variables are low. Then the number of factors necessary to reproduce the variance of primary variables may be greater than half the number of primary variables (see subsection \ref{examp4}).
	\end{itemize}
	The results of the analysis carried out lead to the conclusion that all the criteria discussed above have deficits, and their application does not always lead to the correct determination of the number of factors/components. Due to these deficits in determining the number of factors/components, inconsistencies can arise. However, since these criteria are blind to the variances of single primary variables, their greatest deficit is the inability to reproduce most of the variances of single primary variables. Both the selected principal components in the principal components analysis and the selected factors in the factor analysis may not sufficiently reproduce the variance of some individual primary variables.
	In response to the above deficits, a new criterion for determining the number of factors in factor analysis was analyzed. This criterion makes it possible to present most of the variances of each of the analyzed primary variables. To enable the application of this criterion, an efficient algorithm for determining the number of factors has been proposed.
	
	The answer to the deficits presented above is the criterion which is also discussed in this article. With regard to this criterion, an algorithm is proposed in section \ref{selectAll} that allows the number of factors to be determined in the factor analysis in such a way that the factor model can represent most of the variance of each of the primary variables.
	On the other hand, it should be emphasized that:
	\begin{itemize}
		\item 	The matrix of factor loadings is identical to the matrix of correlation coefficients between the original variables and the principal components obtained in the principal components analysis.
		\item 	The algorithm for estimating factor loadings has a lower time and memory complexity than the algorithm for estimating correlation coefficients between primary and principal variables.
	\end{itemize}
	Therefore, the algorithm for determining the number of factors in factor analysis can also be used to determine the number of principal components in principal component analysis.
	As a result, the number of factors/components can be effectively determined so that most of the variance of each of the primary variables can be represented, not just their mean variance:
	\begin{itemize}
		\item 	In factor analysis, the algorithm selects a sufficient number of factors so that the factor model reproduces most of the variance for each of the primary variables.
		\item 	In principal components analysis, the algorithm selects enough principal components to represent most of the variance of each of the primary variables\footnote{Principal components analysis can be viewed as lossy compression, where several principal components carry most of the information contained in the primary variables. Common sense says that lossy compression assumes that most of the information for all primary variables can be reconstructed from a compressed dataset. An unsatisfactory reconstruction of any primary variable would not achieve this lossy compression goal.}.
	\end{itemize}
	\subsubsection{A necessary condition to determine the optimal number of fac\-tors/com\-po\-nents}
	Due to the need to represent most of the variances of individual primary variables, the criterion presented here can be considered a necessary condition for the correct solution of the problem of determining the number of factors/components.
	It also seems that this criterion can be considered a sufficient condition for a reasonable determination of the number of principal components in principal component analysis.
	
	Unfortunately, this cannot be a sufficient condition for factor analysis. The section \ref{vectFA} describes a case in which it was shown that while only three factors are sufficient to represent most of the variances of all primary variables, only four factors (after Varimax rotation) have made it possible to associate most of the variances of individual primary variables with single factors. It means that only for four factors it was possible to clustered primary variables due to their similarity to the factors.
	\subsubsection{Artifact}
	Subsection \ref{artefakt} describes the observed phenomenon (similarly to\cite{Gniazdowski2017}) in which the product of the matrix $ L $ by the matrix $ U^T $ results in a symmetric matrix:
	\begin{equation}\label{rArtifact1}
		L\cdot U^T=\left(L\cdot U^T\right)^T.
	\end{equation}
	The article \cite{Gniazdowski2017} asks questions about the cause of the observed phenomenon, as well as its potential application. Although there is still no answer to the second question, there is an answer to the first question in the area of factor analysis. Since the matrix of factor loadings in FA is identical to the matrix of correlation coefficients between primary variables and principal components in PCA, this answer is also valid in PCA.
	
	Formula (\ref{r27}) presents $ L $ as the product of the matrix $ U $ and the diagonal matrix $ S $. The diagonal matrix $ S $ can be expressed as the square of two diagonal matrices $ D=\sqrt{S} $
	\begin{equation}\label{rArtifact2}
		S=D\cdot D.
	\end{equation}
	Hence:
	\begin{equation}\label{rArtifact3}
		L\cdot U^T=U\cdot D\cdot D\cdot U^T.
	\end{equation}
	Using the association law for the matrix product, the right side of the above expression can be grouped. Then the expression (\ref{rArtifact3}) takes the form:
	\begin{equation}\label{rArtifact4}
		L\cdot U^T=\left(U\cdot D\right)\cdot\left(D\cdot U^T\right).
	\end{equation}
	Since the transposition of a diagonal matrix is the same matrix, therefore the expression (\ref{rArtifact4}) can be expressed as follows:
	\begin{equation}\label{rArtifact5}
		L\cdot U^T=\left(U\cdot D\right)\cdot\left(U\cdot D\right)^T.
	\end{equation}
	The right side of the expression (\ref{rArtifact5}) shows the product of the matrix by its transposition, so the product $ L\cdot U^T $ is symmetrical.
	\subsubsection{Clustering of random variables}
	The article \cite{Gniazdowski2018} presents a wide spectrum of problems related to clustering of primary variables. For this purpose, various methods of defining dissimilarity of random variables (Euclidean metric, cosine measure) were used, as well as various clustering algorithms (k-means algorithm, spectral algorithms). In particular, the correlated primary random variables have been clustered due to the degree of their similarity to the principal components as well as their similarity to one another.
	
	One of the goals of factor analysis is to find the similarity of the primary variables to the identified interpretable factors. This similarity finding is equivalent to clustering a set of primary variables. The rotation of the identified factors can help in this.
	
	On the other hand, since the matrix of correlation coefficients between primary variables and principal components is identical to the matrix of factor loadings, clustering of primary variables due to their similarity to principal components (described in \cite{Gniazdowski2018}) is the same as clustering, which uses matrix of factor loadings before their rotation.
	
	In the above context, several facts should be noted:
	\begin{itemize}
		\item The algorithm for estimating the factor loadings has a much lower computational complexity (both time and memory) than the algorithm for estimating the matrix of correlation coefficients between primary variables and principal components.
		\item 	The factor loadings matrix enables the representation of primary variables in vector form as points in the space of the Cartesian coordinate system. Clustering of primary variables, due to their similarity to factors/components, refer to the vector representation.
		\item 	The efficiency of the clustering algorithms used in \cite{Gniazdowski2018} does not depend on the rotation of the coordinate system.
	\end{itemize}
	One of the intentions of factor analysis is to clustered primary variables due to their similarity to independent factors. Due to the facts presented above, it is reasonable to conclude that, regardless of the type of analysis (FA or PCA), clustering of primary variables due to their similarity to factors/components should be performed only with the use of the factor loadings matrix.
	On the other hand, with regard to clustering of primary variables, due to their similarity to factors, it seems correct to conclude that clustering of primary variables should not depend on factor rotation.
	
	Before rotation, the factors are equivalent to the standardized principal component.
	Rotation finds factors other than standardized principal components.
	It can be assumed that in the case of a simple analysis of the factor loadings matrix (without the use of a computer), rotation will only facilitate clustering. It is assumed that these new factors will allow for easier grouping of primary variables. This hypothesis should possibly be tested in further research.
	
	\section{Conclusions}
	The article discusses selected problems related to both principal component analysis (PCA) and factor analysis (FA). In particular, both types of analysis were compared. The comparison was limited to principal components analysis, which uses a matrix of correlation coefficients instead of a covariance matrix. Factor analysis was limited to exploratory factor analysis, which uses principal components. 
	
	Comparing principal component analysis and factor analysis not only confirms the existence of many common elements in both types of analysis, but above all reveals three important facts:
	\begin{itemize}
		\item The matrix of factor loadings is identical to the matrix of correlation coefficients between primary variables and principal components obtained in principal components analysis.
		\item The algorithm for estimating the factor loadings has a lower time and memory complexity than the algorithm for estimating the correlation coefficients between primary variables and principal components.
		\item There is a vector interpretation of primary variables. In this interpretation, the respective factor loadings are the components of the vector that represents the given primary variable.
	\end{itemize}
	Therefore, all operations performed on factors/components (determining the number of factors/components) and on primary variables (clustering) can be performed on the factor loadings matrix, and the vector interpretation of primary variables leads to useful conclusions and gives real possibilities of its use:
	\begin{itemize}
		\item The Pythagorean theorem can be used to describe the behavior of primary variables.
		\item 	The cosines between the vectors representing the primary variables are identical to the correlation coefficients between the corresponding primary variables.
		\item 	Based on the vector representation of the primary variables, the number of  fac\-tors/com\-po\-nents can be determined so that they can represent most of the variances of all the primary variables. For this purpose, an appropriate algorithm has been proposed.
		\item 	The condition for the number of factors/components, which enables the representation of most of the variance of each of the primary variables, is a necessary and sufficient condition to determine the optimal number of principal components in principal components analysis and a necessary condition to determine the optimal number of factors in factor analysis.
		\item 	Reducing the number of factors/components is the same as reducing the size of the space in which the primary variables are represented.
		\item 	Based on the Pythagorean theorem, it is possible to analyze the standard deviations and variance of individual original variables by analyzing the lengths and squares of the lengths of the respective vector components.
		\item 	Clustering of primary variables due to their mutual similarity, and also due to the similarity to factors in factor analysis, and also due to the similarity to principal components in principal component analysis, can be performed by clustering vectors (points) due to their mutual similarity and because of the similarity to the factors/components.
	\end{itemize}
	In addition to the practical aspects considered in the article, it is also worth noting an aspect that probably has no practical significance, but is somewhat surprising. This is true for the artifact that has been observed for the vector representation of the primary variables in both PCA and FA. By multiplying the matrix of row vectors representing the primary variables by the transposition of the matrix of eigenvectors $ U^T $, a symmetric matrix was obtained. In this context, questions arose about the cause of the phenomenon and about the possibility of its use. There is no answer to the second question. The article found an algebraic answer to the first question.
	
	\section*{Acknowledgments}
	The author expresses his gratitude to Andrzej Ptasznik for making the Weather Data available for analysis.
	
	\bibliography{biblio}\label{bibliography}
	\bibliographystyle{IEEEtran}
\end{document}